\documentclass[acmtog]{acmart}

\usepackage[linesnumbered,ruled,vlined]{algorithm2e}
\usepackage{booktabs} 
\usepackage{pifont}
\usepackage{lipsum}

\newcommand\blfootnote[1]{%
  \begingroup
  \renewcommand\thefootnote{}\footnote{#1}%
  \addtocounter{footnote}{-1}%
  \endgroup
}

\usepackage[ruled,vlined]{algorithm2e}
\newcommand{\cmark}{\ding{51}}%
\newcommand{\xmark}{\ding{55}}%
\SetAlCapFnt{\small}
\SetAlCapNameFnt{\small}
\SetAlCapHSkip{0pt}

\acmJournal{TOG}
\newcommand{\revise}[1]{\textcolor{black}{#1}}
\usepackage{ulem}
\usepackage{xcolor}
\usepackage{multirow}
\usepackage{makecell}
\begin{document}

\acmJournal{TOG}
\acmYear{2024} \acmVolume{43} \acmNumber{6} \acmArticle{} \acmMonth{12}\acmDOI{10.1145/3687980}

\title{Human4DiT: \revise{360-Degree}  Human Video Generation with 4D Diffusion Transformer}

\author{Ruizhi Shao*}
\orcid{}
\affiliation{%
 \institution{Tsinghua University}
 \department{Department of Automation}
 \city{Beijing}
 \country{China}
}

\author{Youxin Pang*}
\orcid{}
\affiliation{%
 \institution{Tsinghua University}
 \department{Department of Automation}
 \city{Beijing}
 \country{China}
}

\author{Zerong Zheng}
\orcid{}
\affiliation{%
 \institution{Tsinghua University}
 \department{Department of Automation}
 \city{Beijing}
 \country{China}
}

\author{Jingxiang Sun}
\orcid{}
\affiliation{%
 \institution{Tsinghua University}
 \department{Department of Automation}
 \city{Beijing}
 \country{China}
}

\author{Yebin Liu}
\orcid{}
\affiliation{%
 \institution{Tsinghua University}
 \department{Department of Automation}
 \city{Beijing}
 \country{China}
}

\begin{abstract}
We present a novel approach for generating 360-degree high-quality, spatio-temporally coherent human videos from a single image. Our framework combines the strengths of diffusion transformers for capturing global correlations across viewpoints and time, and CNNs for accurate condition injection. The core is a hierarchical 4D transformer architecture that factorizes self-attention across views, time steps, and spatial dimensions, enabling efficient modeling of the 4D space. Precise conditioning is achieved by injecting human identity, camera parameters, and temporal signals into the respective transformers. To train this model, we collect a multi-dimensional dataset spanning images, videos, multi-view data, and limited 4D footage, along with a tailored multi-dimensional training strategy. Our approach overcomes the limitations of previous methods based on generative adversarial networks or vanilla diffusion models, which struggle with complex motions, viewpoint changes, and generalization. Through extensive experiments, we demonstrate our method's ability to synthesize 360-degree realistic, coherent human motion videos, paving the way for advanced multimedia applications in areas such as virtual reality and animation.
\end{abstract}

\begin{teaserfigure}
    \captionsetup{type=figure}
    \includegraphics[width=1.\textwidth]{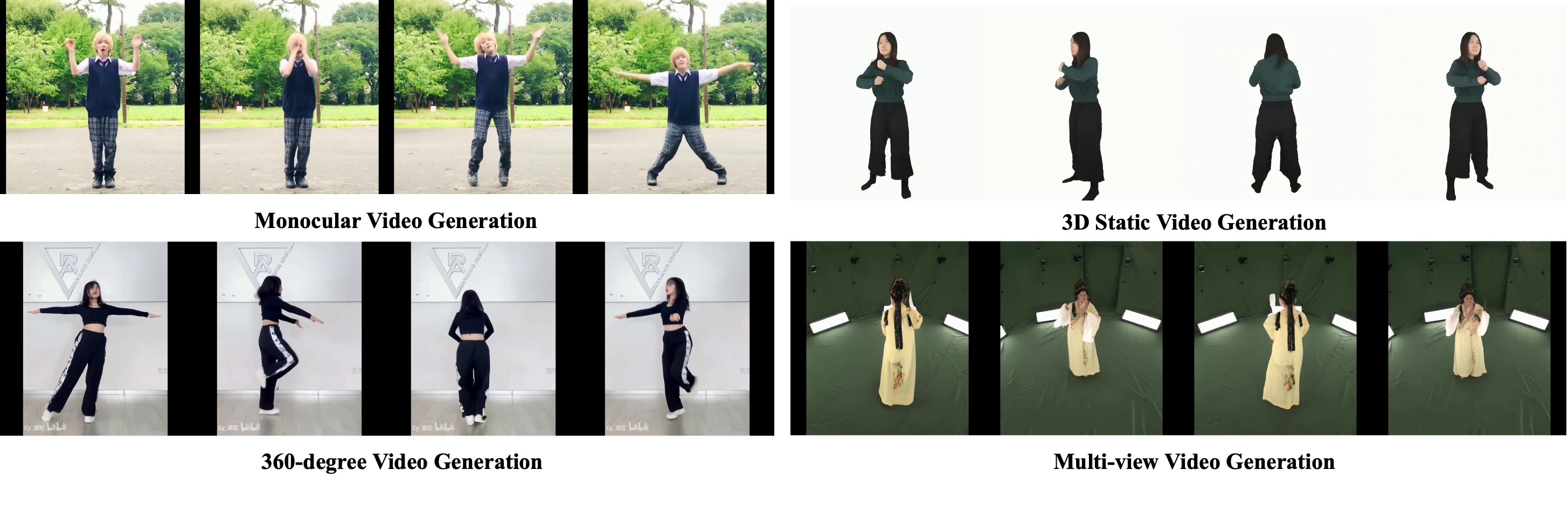}
    \vspace{-8mm}
    \captionof{figure}{We propose \textbf{Human4DiT}, a novel approach to generate 360-degree high-quality, spatio-temporally coherent human videos given a reference image. With the proposed 4D diffusion transformer, our method is capable of generating monocular video, multi-view video, 3D static video, and 360-degree rotating video.
    }
\end{teaserfigure}

\maketitle

\blfootnote{* equal contribution.}
\section{Introduction}
Human video generation is an active research topic in the field of video generation. It has widespread applications in areas such as virtual reality, animation, gaming, and movie production. Moreover, generating realistic human videos holds great significance for advancing multimedia technologies and enabling new forms of human-computer interaction.

Recently, with the rapid development of diffusion models, especially latent diffusion models~\cite{rombach2022high}, leveraging diffusion models for human video generation has become the mainstream approach~\cite{wang2023disco, xu2023magicanimate, zhu2024champ, hu2023animate}. To incorporate human priors as control conditions into the diffusion model, some approaches~\cite{lee2023same} have adopted skeleton-based schemes, using the skeletal connectivity graph as a control condition either through ControlNet~\cite{zhang2023adding} or by concatenating it with the input. Other methods are based on the SMPL body model template~\cite{loper2023smpl}, injecting SMPL-derived representations such as UV maps, depth, normals~\cite{zhu2024champ}, or direct dense pose embeddings~\cite{xu2023magicanimate} into the diffusion model. Current human video diffusion models based on the CNNs architecture~\cite{ronneberger2015u} such as UNet from Stable Diffusion could inject control conditions into the network in a pixel-aligned manner. However, the UNet's reliance on local convolutional operations makes it more focus on local generation, resulting in relatively poorer performance for the global aspects, especially when generating long and complex human motions. 
Moreover, these methods only consider the human body itself, neglecting viewpoint information from the camera perspective, especially for scenarios involving significant viewpoint changes such as 360-degree human video generation. Incorporating viewpoint control signals into the network while simultaneously maintaining coherence across different viewpoints and time poses a significant challenge.

To overcome the challenges of generating complex human motions across views and time, we propose a novel video generation network architecture that combines the strengths of CNNs and diffusion transformers. First, we propose to leverage 3D SMPL models instead of 2D skeleton maps to efficiently incorporate view information and correspondences across multiple viewpoints. Our method employs the rendering of normal human maps as robust view-dependent guidance. These rendered normal maps are subsequently processed through a CNN-based encoder, which encourages the network to capture view-dependent information and precisely inject pixel-aligned conditions.

However, it remains a challenge to ensure temporal consistency when using CNNs-based architectures to generate long videos. 
Recently, OpenAI's recent work on long video generation, SORA~\cite{sora2024}, adopted a diffusion transformer~\cite{peebles2023scalable} architecture and demonstrated substantially better realism and spatio-temporal coherence than CNNs-based models. Inspired by SORA, we introduce a 4D diffusion transformer for human video generation, which not only exhibits greater potential for scalability, but also demonstrates the capacity to learn complex 360-degree human video generation. By employing the unified attention mechanism across various dimensions, our 4D DiT could efficiently build spatial-temporal correspondences across different views and time, preserving spatial-temporal consistency in the generated human videos.
However, directly applying a diffusion transformer to learn correlations over the views and time would be computationally prohibitive. Therefore, we propose an efficient novel 4D transformer architecture. Its core principle is to cascadingly learn the correlations across the 4D space (view, time, height, width) via self-attention. Specifically, we factorize the 4D diffusion transformer into three transformer blocks: 2D image transformer blocks, temporal transformer blocks, and view transformer blocks, each attending to different dimensions of the 4D space. These three types of blocks are interconnected to form a 4D transformer block. Multiple such 4D transformer blocks are then cascaded to construct the final 4D transformer.
This efficiently captures the interrelations between body parts (height, width) across viewpoints (views) and time steps (times). 

To enhance the controllability of our proposed 4D diffusion transformer beyond SMPL motion, we integrate additional control signals into the respective network modules, including human identity, temporal information, and camera parameters. Human identity embeddings and latent tokens, extracted via the CLIP and  a CNNs-based encoder, are incorporated into image transformers. Camera embeddings, derived from camera parameters, are integrated into the view transformer, while temporal embeddings are incorporated into the temporal transformer. Through these modules, we effectively inject various control conditions into the network, facilitating viewpoint manipulation and the generation of high-fidelity, consistent human videos.

To train our proposed 4D diffusion transformer model, we also collected a large multi-dimensional dataset and devised a multi-dimensional training strategy that fully leverages all available data modalities. Our multi-dimensional dataset comprises images, videos, multi-view videos, 3D scans, as well as a limited amount of 4D scans spanning different viewpoints and time steps. 
During the inference stage, we propose a spatio-temporal consistent diffusion sampling strategy that enables the generation of long 360-degree videos despite limited spatio-temporal window constraints. The strategy is carried out in two stages. The first stage treats the 360-degree video as a monocular long video sequence, maximizing the temporal window to ensure long-term temporal consistency. The second stage regards the 360-degree video as a collection of multi-view video clips, using a larger viewpoint window with a smaller temporal window, encouraging consistency across viewpoints.

To summarize, our main contributions are:
\begin{itemize}
\item We are the first to introduce diffusion transformers to human video generation. Combined with methods that simply use CNNs-based encoder and several control modules, our method achieves high-quality 360-degree spatio-temporally consistent generation of long human videos.
\item We propose an efficient 4D diffusion transformer architecture composed of three transformers attending to 2D images, time, and viewpoints respectively, significantly reducing computational requirements while effectively capturing correlations between body parts across space and time.
\item We collect a large multi-dimensional 4D human dataset and introduce a multi-dimensional training strategy that fully leverages data from all modalities. 
\item During inference, we also propose a spatio-temporal consistent diffusion sampling strategy to generate coherent 360-degree long human videos.

\end{itemize}

\begin{figure*}[ht!]
    \centering
    \includegraphics[width=\linewidth]{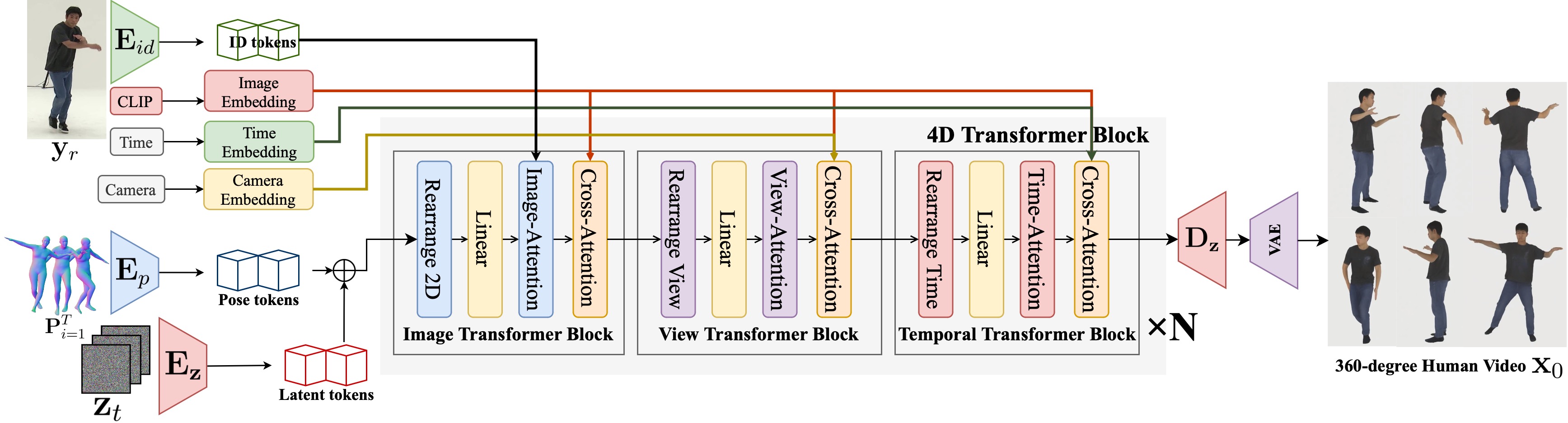}
    \vspace{-3mm}
    \caption{\textbf{\textit{Pipeline of Human4DiT:} our framework is based on 4D diffusion transformer, which adopts a cascaded structure consisting of the 2D image transformer, the view transformer, and the temporal transformer.
    The input contains a reference image $\mathbf{y}_{r}$, dynamic SMPL sequences $\mathbf{P}$, and camera parameters $\mathbf{c}$. 
    Starting from a generated noisy latent representation $\mathbf{z}_{t}$, we denoise them conditioned on $\mathbf{y}_{r}$, $\mathbf{P}$, and $\mathbf{c}$ to recover the original latent frames. 
    First, the 2D image transformer block is designed to capture spatial self-attention within latent frame tokens and pose frame tokens extracted by latent encoder $\mathbf{E}_\mathbf{z}$ and pose encoder $\mathbf{E}_{p}$, respectively. 
    In addition, ID tokens $\mathbf{y}_{id}$ and image embedding $\mathbf{y}_e$ extracted from $\mathbf{y}_{r}$ by ID encoder $\mathbf{E}_{id}$ and CLIP are also injected to ensure identity consistency.
    Secondly, we use the view transformer block to learn correspondences across different viewpoints conditioned on camera embedding.
    Finally, we adopt a temporal transformer to capture temporal correlations with time embedding. 
    The time embedding and camera embedding are obtained by positional encoding time $\mathbf{T}$ and camera $\mathbf{c}$, respectively.$^5$}}
    \label{fig:pipeline}
\end{figure*}

\section{Related Work}
\subsection{Human Video Generation}
Human video generation is the task of creating realistic and temporally coherent videos of humans from input data such as text descriptions, images, or motion sequences. The current paradiam of human video generation lies in two main categories: GAN-based and diffusion-based approaches. 

GAN-based methods \cite{siarohin2019first, tian2021good, wang2021one, wang2020g3an, siarohin2018deformable, siarohin2019appearance} leverage the inherent generative capabilities of adversarial networks \cite{goodfellow2014generative, mirza2014conditional} to spatially transform reference images according to input motion. These approaches commonly employ warping functions to generate sequential video frames, aiming to fill in missing regions and improve visually implausible areas within the generated content. While GAN-based methods have shown promising results in dynamic visual content generation, they often struggle with effectively transferring motion, especially when there are significant variations in human identity and scene dynamics between the reference image and the source video motion. This can lead to unrealistic visual artifacts and temporal inconsistencies in the synthesized videos.

On the other hand, diffusion models, known for the superior generation quality and stable controllability, have been successfully integrated into human image animation. These models~\cite{bhunia2023person, karras2023dreampose, wang2023disco, hu2023animate, xu2023magicanimate, zhu2024champ} employ various strategies, such as texture diffusion blocks, optical flow synthesis in latent space, and motion representation using flow maps, to enhance the visual fidelity of the generated videos. Animate Anyone \cite{hu2023animate} employs a UNet-based ReferenceNet to extract features from reference images and incorporates pose information through an efficient pose guider. However, recent diffusion-based methods are still mainly relying on image diffusion model and face challenges in maintaining texture consistency and temporal stability across frames. Furthermore, these methods don't explore view controllability.

\subsection{Diffusion Model with Camera Control}
Camera information is conditioned in diffusion models for view control. One line of works inject camera parameters into the text-to-image diffusion models for consistent view synthesis from a single input image, which has inspired researchers to explore their potential for generating consistent multi-view images~\cite{shi2023mvdream, liu2023zero, kuang2024collaborative, cai2023genren}. Zero-1-to-3~\cite{liu2023zero} finetunes Stable diffusion by conditioning camera poses for zero-shot novel view synthesis. Syncdreamer~\cite{liu2023syncdreamer} employs 3D volumes and depth-wise attention to maintain consistency across views. MVDream~\cite{shi2023mvdream} and other methods like Wonder3D~\cite{long2023wonder3d} and Zero123++~\cite{shi2023zero123++} leverage 3D self-attention to extend multi-view image generation to more general and efficient. There are other works~\cite{yang2024direct, he2024cameractrl} inject camera information into text-to-video (T2V) models. For example, Direct-a-Video~\cite{yang2024direct} injects quantitative camera movements into temporal cross-attention blocks for view control. CameraCtrl~\cite{he2024cameractrl} proposes to use an efficient representation, Plücker ray embedding, for camera conditions.

\subsection{Diffusion Transformer}
The transformer architecture~\cite{vaswani2017attention} has revolutionized the field of natural language processing, with models like GPT~\cite{radford2018improving, radford2019language} achieving remarkable success. Recent research has demonstrated the potential of transformers in various computer vision tasks, including image classification~\cite{touvron2021training, yuan2021tokens}, semantic segmentation~\cite{zheng2021rethinking, xie2021segformer, strudel2021segmenter}. 

Building upon this progress, the Diffusion Transformer (DiT)~\cite{peebles2023scalable} and its variants~\cite{bao2023all, zheng2023fast} have taken a step further by substituting the conventional convolutional-based U-Net backbone~\cite{ronneberger2015u} with transformers in diffusion models. This architectural shift offers enhanced scalability compared to U-Net-based models, enabling the seamless expansion of model parameters. Recently, the transformer architecture is also integrated into text-to-video models~\cite{lu2023vdt, sora2024, ma2024latte}, improving generation performance.

\section{Overview}
Given a reference image of a person $\mathbf{y}_{r}$, a sequence of dynamic SMPL models $\{\mathbf{P}_{i=1}^T=\mathbf{(\theta, \beta)}_{i=1}^T\}$, and camera parameters $\{\mathbf{c}_{i=1}^V\}$, the goal of our method is to generate a video of that person performing the corresponding motion from the specified viewpoint. The overall pipeline of our approach is illustrated in Figure~\ref{fig:pipeline}. Since our framework is based on a latent diffusion transformer, we first generate a noisy latent representation $\mathbf{z}_t$. We then inject $\mathbf{y}_{r}$, $\{\mathbf{P}_{i=1}^T\}$, and $\{\mathbf{c}_{i=1}^V\}$ as control signals into the diffusion transformer, which iteratively performs denoising to ultimately generate the latent video $\mathbf{z}_0$ and decoded 360-degree human video $\mathbf{x}_0$(Sec.~\ref{section:network_structure}).

To train our proposed model, we collected a multi-dimensional human dataset comprising images, videos, multi-view data, as well as 3D and 4D human data. We further introduce a multi-dimensional mixed training strategy that fully leverages all available data modalities for effective network training (Sec.~\ref{section:dataset}).

During the inference stage, to enable the generation of 360-degree long-duration human motion videos under limited temporal and viewpoint window constraints, we propose an efficient diffusion sampling strategy. This strategy achieves improved spatial-temporal coherence by two-staged viewpoint and temporal window planning in the diffusion sampling process (Sec.~\ref{section:sampling}).

\begin{figure}
     \centering
    \includegraphics[width=\linewidth]{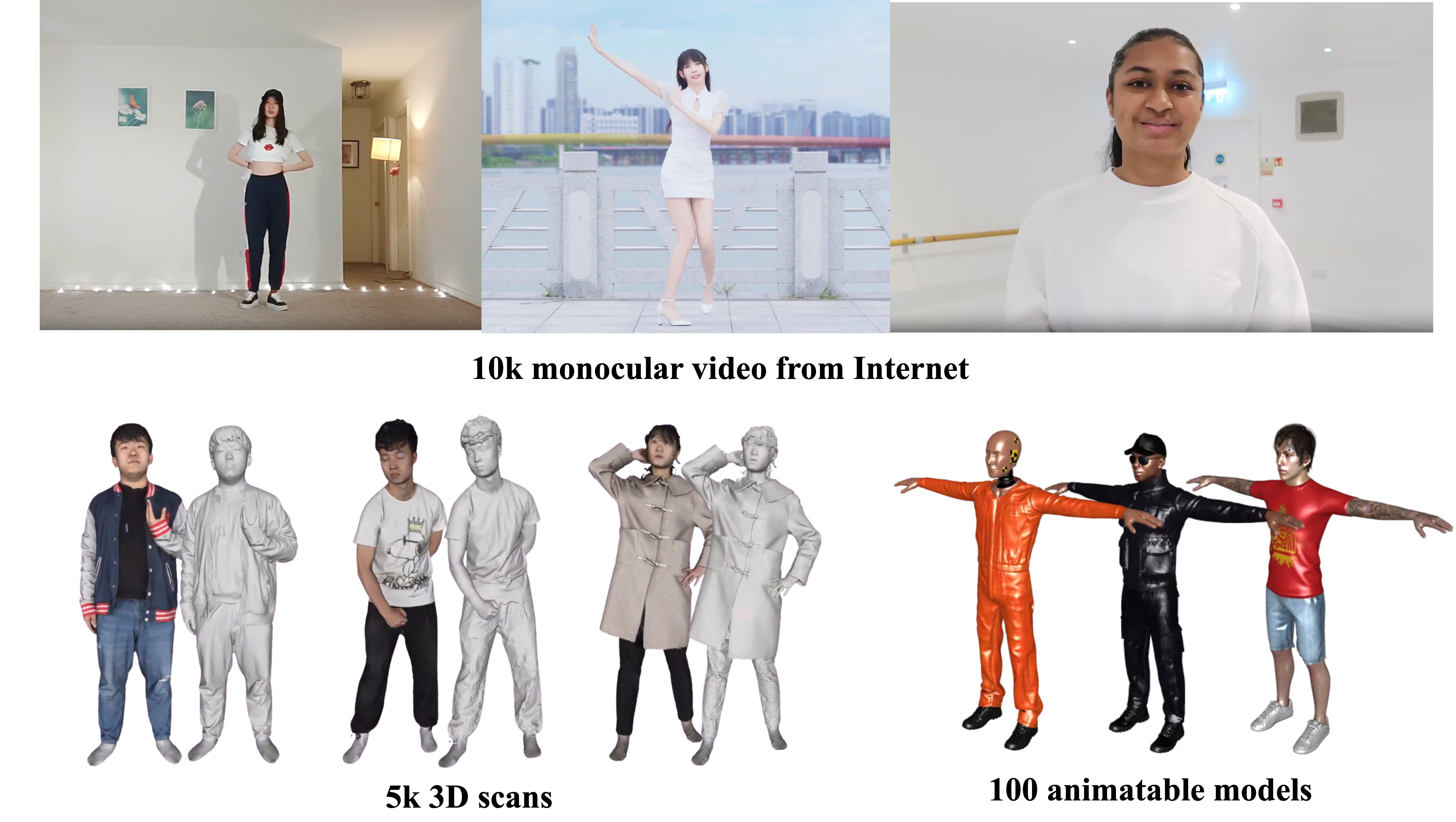}
    \vspace{-3mm}
    \caption{\textbf{\textit{Human4DiT dataset}}: In addition to the open-source dataset, we collect a multi-dimension dataset including 10k monocular videos from Internet, 5k high-quality 3D human scans and 100 animatable human models for dynamic free-view rendering. }
    \label{fig:human4dit_dataset}
\end{figure}

\section{Network Structure}
\label{section:network_structure}
In this section, we introduce the core components of our network structure: the 4D diffusion transformer and the control condition injection modules. To efficiently establish spatial-temporal relationships, our 4D diffusion transformer adopts a cascaded structure consisting of the 2D image, the temporal, and the view transformer blocks. Additionally, our control condition injection modules comprise the reference image injection module, the SMPL injection module, the time injection module, and the viewpoint injection module. Notably, the reference image injection module and the SMPL injection module utilize CNNs-based architectures to ensure pixel-aligned injection of the control conditions.

\begin{table*}[ht!]
\small
    \centering
    \begin{tabular}{c|cccccccc}
    \hline
        Dataset & Clips & Frames & Views & Resolution & 3D & Dynamic & SMPL & Body \\ \hline
        HumanArt~\cite{ju2023human} & - & 50k & 1 & $512^2$-$1024^2$ & \xmark & \xmark & Ground Truth & Half-body + Full-body \\ 
        TikTok~\cite{jafarian2021learning} & 2k & 600k & 1 & 480P-720P & \xmark & \cmark & Fitting & Half-body + Full-body \\ 
        TalkShow~\cite{yi2023generating} & 1k & 300k & 1 & 480P-720P & \xmark & \cmark & Fitting & Half-body \\ 
        AIST~\cite{aist-dance-db} & 10k & 2000k & 6 & 1080P & \xmark & \cmark & Ground Truth & Full-body \\ 
        Motion-X~\cite{lin2024motion} & 10k & 1000k & 1 & 720P-1080P & \xmark & \cmark & Ground Truth & Full-body \\ 
        DNA-rendering~\cite{cheng2023dna} & 2k & 400k & 16 & 4K & \xmark & \cmark & Fitting & Full-body \\ 
        Twindom~\cite{twindom} & 2k & 360k & 180 & $1024^2$ & \xmark & \cmark & Fitting & Full-body \\
        THuman2.0~\cite{tao2021function4d} & 500 & 90k & 180 & $1024^2$ & \cmark & \xmark & Ground Truth & Full-body \\
        THuman-CloSET~\cite{zhang2023closet} & 500 & 90k & 180 & $1024^2$ & \cmark & \xmark & Ground Truth & Full-body \\
        Bedlam~\cite{black2023bedlam} & 10k & 1500k & 1 & 720P-1080P & \xmark & \cmark & Ground Truth & Full-body \\ \hline
        \textbf{Human4DiT-3D} & 5k & 900k & 180 & $1024^2$ & \cmark & \xmark & Fitting & Full-body \\
        \textbf{Human4DiT-Video} & 10k & 2000k & 1 & 720P-1080P & \xmark & \cmark & Fitting & Half-body + Full-body \\
        \textbf{Human4DiT-4D} & 100 & 168k & 180 & $1024^2$ & \cmark & \cmark & Fitting & Full-body \\
         \hline
    \end{tabular}
    \caption{The collected multi-dimensional training dataset.}
    \label{tab:dataset}
    \vspace{-6mm}
\end{table*}

\subsection{4D Diffusion Transformer}
Our 4D diffusion transformer performs denoising on a randomly initialized noisy multi-view latent human video $\mathbf{z}_t$. After multiple denoising steps, it produces the denoised output $\mathbf{z}_0$, which is eventually decoded by a VAE to generate the video. We first utilize a CNNs-based encoder $\mathbf{E}_\mathbf{z}$ to extract latent tokens. Then the input could be regarded as a set of tokens in 5D ($\mathbf{z}_t \in \mathbb{R}^{V \times T \times H \times W \times C}$), where $C$ represents the feature dimension of each token, $V$ is the viewpoint dimension, $T$ is the time dimension, and $H$ and $W$ are the height and width, respectively. To establish correlations across viewpoints, time, and 2D spatial dimensions, we propose a novel cascaded diffusion transformer architecture. It could decompose the complex 4D attention into three types of attention including image, temporal and view attention, which efficiently learns the 4D correspondences and decreases the memory usage. 

First, we feed $z_t$ into a 2D image transformer block to capture spatial self-attention within each frame:

\begin{equation}
    \begin{split}
\mathbf{z}_t^s &= \text{rearrange}(\mathbf{z}_t, (V \times T, H \times W, C)), \\
\mathbf{Q}_s, \mathbf{K}_s, \mathbf{V}_s & = f_s^\mathbf{Q}(\mathbf{z}_t^s), f_s^\mathbf{K}(\mathbf{z}_t^s), f_s^\mathbf{V}(\mathbf{z}_t^s), \\
\mathbf{\hat{z}}_t^s & = \text{softmax}(\frac{\mathbf{Q}_s\mathbf{K}_s^T}{\sqrt{d_k}})\mathbf{V}_s,
    \end{split}
\end{equation}
where $f_s^\mathbf{Q}, f_s^\mathbf{K}, f_s^\mathbf{V}$ are the linear layers in the transformer block. The rearrange operation reshapes the tensor $\mathbf{z}_t$ into a 3D tensor $\mathbf{z}_t^s \in R^{(V \times T) \times (H \times W) \times C}$, where the first dimension $V \times T$ is treated as the batch size. The actual attention operation is then performed on the last two dimensions $H \times W$ and $C$, which correspond to the 2D spatial dimensions of the images. After the 2D image transformer, we further model correspondences across different viewpoints through the view transformer block:
\begin{equation}
    \begin{split}
\mathbf{z}_t^v &= \text{rearrange}(\mathbf{\hat{z}}_t^s, (T, V \times H \times W, C)), \\
\mathbf{Q}_v, \mathbf{K}_v, \mathbf{V}_v & = f_v^\mathbf{Q}(\mathbf{z}_t^v), f_v^\mathbf{K}(\mathbf{z}_t^v), f_v^\mathbf{V}(\mathbf{z}_t^v), \\
\mathbf{\hat{z}}_t^v & = \text{softmax}(\frac{\mathbf{Q}_v\mathbf{K}_v^T}{\sqrt{d_k}})\mathbf{V}_v,
    \end{split}
\end{equation}

In the view transformer, due to the substantial variations across different viewpoints, especially between frontal, side, and back views, we perform attention jointly across the viewpoint and the 2D spatial dimensions. This attention allows for better modeling of the global correlations across the entire frame. It also makes the view transformer the most computationally and memory-intensive component, limiting the maximum viewpoint window size. We will address this constraint during the inference process, as discussed later.
After the view transformer, we finally employ a temporal transformer to capture temporal correlations across time steps. 

\begin{equation}
    \begin{split}
\mathbf{z}_t^m &= \text{rearrange}(\mathbf{\hat{z}}_t^v, ( V \times H \times W, T, C)), \\
\mathbf{Q}_m, \mathbf{K}_m, \mathbf{V}_m & = f_m^\mathbf{Q}(\mathbf{x}_t^m), f_m^\mathbf{K}(\mathbf{z}_t^m), f_m^\mathbf{V}(\mathbf{z}_t^m), \\
\mathbf{\hat{z}}_t^m & = \text{softmax}(\frac{\mathbf{Q}_m\mathbf{K}_m^T}{\sqrt{d_k}})\mathbf{V}_m,
    \end{split}
\end{equation}
The three transformer blocks (2D image, view, and temporal) are interconnected to form a single 4D transformer block. Our complete 4D diffusion transformer architecture is composed of 10 cascaded 4D transformer blocks.
Through this cascaded multi-level attention scheme, our approach significantly reduces the computational overhead while improving the training efficiency and effectively ensuring spatio-temporal consistency in the generation process.

\subsection{Control Modules}
\subsubsection{Camera Control Module}
To incorporate camera viewpoint control into the network, we assume the first camera $\mathbf{c}_1$ as the world coordinate system, extract its rotation matrix (i.e., the identity matrix) as a 9D tensor $\mathbf{r}_1$, and apply positional encoding:
\begin{equation}
\mathbf{y}_c = (\sin(2^0\pi\mathbf{r}_1), \cos(2^0\pi\mathbf{r}_1),...,(\sin(2^{L-1}\pi\mathbf{r}_1), \cos(2^{L-1}\pi\mathbf{r}_1))
\end{equation}
The parameters of other cameras $\{\mathbf{c}_{i=2}^V\}$ are computed as rotation matrices relative to the first camera, and their encodings are obtained through the same positional encoding process. We then map these encodings to the same dimension as the CLIP image embeddings using an MLP $f_c$ and inject them into the view transformer module via addition to influence the intermediate features after self-attention:
\begin{equation}
\mathbf{z}^{v'}_t = \mathbf{z}^v_t + f_c(\mathbf{y}_c)
\end{equation}
This relative encoding formulation effectively represents the correlations between different viewpoints, enabling better generation with varying multi-view setups.

\subsubsection{Temporal Embedding Module}
To incorporate temporal control into the network, we apply positional encoding directly to the time $T_m$ (frame number):
\begin{equation}
\mathbf{y}_m = (\sin(2^0\pi T_m), \cos(2^0\pi T_m),...,(\sin(2^{L-1}\pi T_m), \cos(2^{L-1}\pi T_m)
\end{equation}
We then map the temporal encoding to the latent space using an MLP $f_m$ and add it to the temporal transformer's features after self-attention:
\begin{equation}
\mathbf{z}^{m'}_t = \mathbf{z}^m_t + f_m(\mathbf{y}_m)
\end{equation}

\subsubsection{SMPL Control Module}
To inject the SMPL parameters into the network, we first obtain the SMPL mesh vertex positions from the given shape and pose parameters $\mathbf{p}_i = LBS(\theta_i, \beta_i)$ of each frame $i$.
We then render these vertices $\mathbf{p}_i$ into normal maps $\mathbf{M}_n(i, v)$ using camera parameters $\mathbf{c}_v$, as normals provide an effective representation of the 3D human. Since the camera parameters $\mathbf{c}_v$ have already been injected into the network, we decouple the SMPL information from the rendered normal maps by multiplying it with the inverse of the camera rotation matrix $\mathbf{r}_v^{-1}$. We then use a CNNs-based Encoder $\mathbf{E}_p$ to extract features $\mathbf{y}_n$ from these SMPL rendered normal maps $\mathbf{M}_n(i, v)$. These features $\mathbf{y}_n$ are eventually added to the input $\mathbf{x}_t$ of the 4D Diffusion Transformer before being fed into the network. Through this approach, the control condition is injected into the network in a pixel-aligned manner, ensuring consistent and accurate human motion video generation.

\subsubsection{Human Identity Reference Module}
During human video generation, we extract the human identity from a reference image and inject it into the network. We employ two ways to maintain identity consistency. First, similar to the SMPL injection, we use a CNNs-based encoder $E_{id}$ to extract human features $\mathbf{y}_{id}$ from the reference image. These features are then added to the input of the 4D diffusion transformer $\mathbf{x}_t$, allowing the network to better capture the detailed identity characteristics from the reference image.
Additionally, we extract an image embedding $\mathbf{y}_{e}$ using CLIP and inject it into each transformer block via cross-attention after the self-attention mechanism:
\begin{equation}
\begin{split}
    \mathbf{Q}, \mathbf{K}, \mathbf{V} & = f^\mathbf{Q}(\mathbf{\hat{z}}_t), f^\mathbf{K}(\mathbf{y}_{e}), f^\mathbf{V}(\mathbf{y}_{e}), \\
\mathbf{\tilde{z}}_t & = \text{softmax}(\frac{\mathbf{Q}\mathbf{K}^T}{\sqrt{d_k}})\mathbf{V}
    \end{split}
\end{equation}
This ensures that the generated output maintains global consistency with the reference human identity in addition to preserving local details.

\section{Dataset and Training Strategy}
\label{section:dataset}
To train this 4D transformer model, we collected a multi-dimensional dataset and devised a multi-dimensional training strategy that fully leverages all available data modalities. As shown in Tab.~\ref{tab:dataset}, in addition to utilizing publicly available human datasets, we collect a new multi-dimensional dataset comprises human images, videos, multi-view videos, 3D data, as well as a small amount of 4D data spanning different viewpoints and time steps. It contains 5k 3D human scans capture by camera rigs, 10k videos from YouTube/Bilibili, and 100 animatable human models from artists. Some samples are presented in Fig.~\ref{fig:human4dit_dataset}. We adopt two methods to estimate SMPL for our dataset: For multi-view and 3D/4D data, we detect 2D key points and optimize SMPLs to align with the corresponding images~\cite{lightcap2021}. For monocular videos, we directly use Humans-in-4D~\cite{goel2023humansIn4d} to estimate image-aligned SMPLs and cameras.

We employ different training strategies for each data modality. For the 2D images dataset, we only train the 2D transformer, using the CLIP image embedding as the human identity condition. For the single-view video dataset, we train the 2D and temporal transformers, randomly selecting one frame from the video as the reference image. For multi-view videos, we train all transformers simultaneously, randomly choosing one frame from the multi-view data as the reference image. For 3D dataset, we train our transformer with two strategies. multi-view images and train only the 2D and view transformers, randomly selecting one rendered view as the reference. For 4D dataset, we render it into dense-view videos or videos with continuous viewpoint movements and train all transformers concurrently.

This multi-dimensional training approach allows us to effectively leverage diverse data sources, with each modality contributing to different components of the model. The unified 4D transformer architecture seamlessly integrates information from all modalities during training, enabling coherent 360-degree human video generation.

\begin{algorithm}
\small
\caption{Efficient Spatial-Temporal Sampling}
\label{alg:ddpm_sampling}
\KwIn{$T_i$ (number of inference timesteps); $\mathbf{y}$ (Control signals); $T_L$ (number of frames); $M_T^1, M_T^2, M_V^2$ (window size of view and time); $\epsilon_\theta$ (4D diffusion transformer)}
\KwOut{$x_0$ (sampled latent video)}
$\mathbf{x}_t \sim \mathcal{N}(0, I),$ $\epsilon \sim \mathcal{N}(0, I)$ \;
$N^1_T = \frac{T_L}{M_T^1},$ 
$N^2_T = \frac{T_L}{M_T^2}, N^2_V = \frac{T_L}{M_V^2}$ \;
\For{$t \gets T_i$ \KwTo $1$}{
    \For{$m \gets 1$ \KwTo $N^1_T$}{ 
            $slice^1 = \left[mM^1_T:(m+1)M^1_T\right]$\;
            
            $\epsilon^1\left[slice^1\right] = \epsilon_\theta(\mathbf{x}_t\left[slice^1\right], t, \mathbf{y}\left[slice^1\right])$\;
        }
    \For{$m \gets 1$ \KwTo $N^2_T,$ $v \gets 1$ \KwTo $N^2_V$} {
            $slice^2 = \left[v:T_L:N^2_V, mM^2_T:(m+1)M^2_T\right]$\;
            
            $\epsilon^2\left[slice^2\right] = \epsilon_\theta(\mathbf{x}_t\left[slice^2\right], t, \mathbf{y}\left[slice^2\right])$\;
        }
    $\mu_\theta \gets \frac{1}{\sqrt{\alpha_t}} \left( x_t - \frac{\beta_t}{\sqrt{1 - \bar{\alpha}_t}} (\lambda_1\epsilon_1 + \lambda_2\epsilon_2) \right)$ \;
    $x_{t-1} \gets \mu_\theta + \sigma_t \epsilon$
}

\Return{$x_0$}\;
\end{algorithm}

\section{Efficient Spatial-Temporal Sampling}
\label{section:sampling}
Our 4D diffusion transformer is capable of generating multi-view human motion videos. However, it cannot directly generate 360-degree videos where both viewpoint and time vary simultaneously. This limitation arises from computational memory constraints, where the total number of frames $M$ in the network input is bounded by the product of the temporal window size $M_T$ and the view window size $M_V$.
To enable generation of long 360-degree videos, we propose an efficient and spatio-temporally consistent sampling method. During the denoising process, this method employs two strategies. The first strategy treats the 360-degree video as a monocular long video sequence, maximizing the temporal window $M_T^1$ to ensure long-term temporal consistency. The second strategy regards the 360-degree video as a collection of multi-view short video clips, using a larger viewpoint window $M_V^2$ with a smaller temporal window $M_T^2$, focusing on maintaining consistency across viewpoints.
During denoising, the noise predictions from these two strategies are combined using respective weightings $\lambda_1, \lambda_2$. The specific sampling algorithm is outlined in Alg.~\ref{alg:ddpm_sampling}.

This sampling approach combines the advantages of both strategies. Globally, it leverages the view transformer to maintain consistency across large viewpoint separations. For smaller inter-frame motions between adjacent viewpoints, the temporal transformer ensures coherence. Ultimately, this method achieves spatio-temporally consistent generation of long 360-degree human motion videos under limited input window constraints.

\section{Experiment}
In this section, to validate the capabilities of our 4D diffusion transformer, we conducted comprehensive comparisons against current state-of-the-art human video generation methods including Disco~\cite{wang2023disco}, MagicAnimate~\cite{xu2023magicanimate}, AnimateAnyone~\cite{hu2023animate}, and Champ~\cite{zhu2024champ} on monocular video, multi-view video, 3D static video, and 360-degree video generation.

\begin{figure*}
    \centering
    \includegraphics[width=\linewidth]{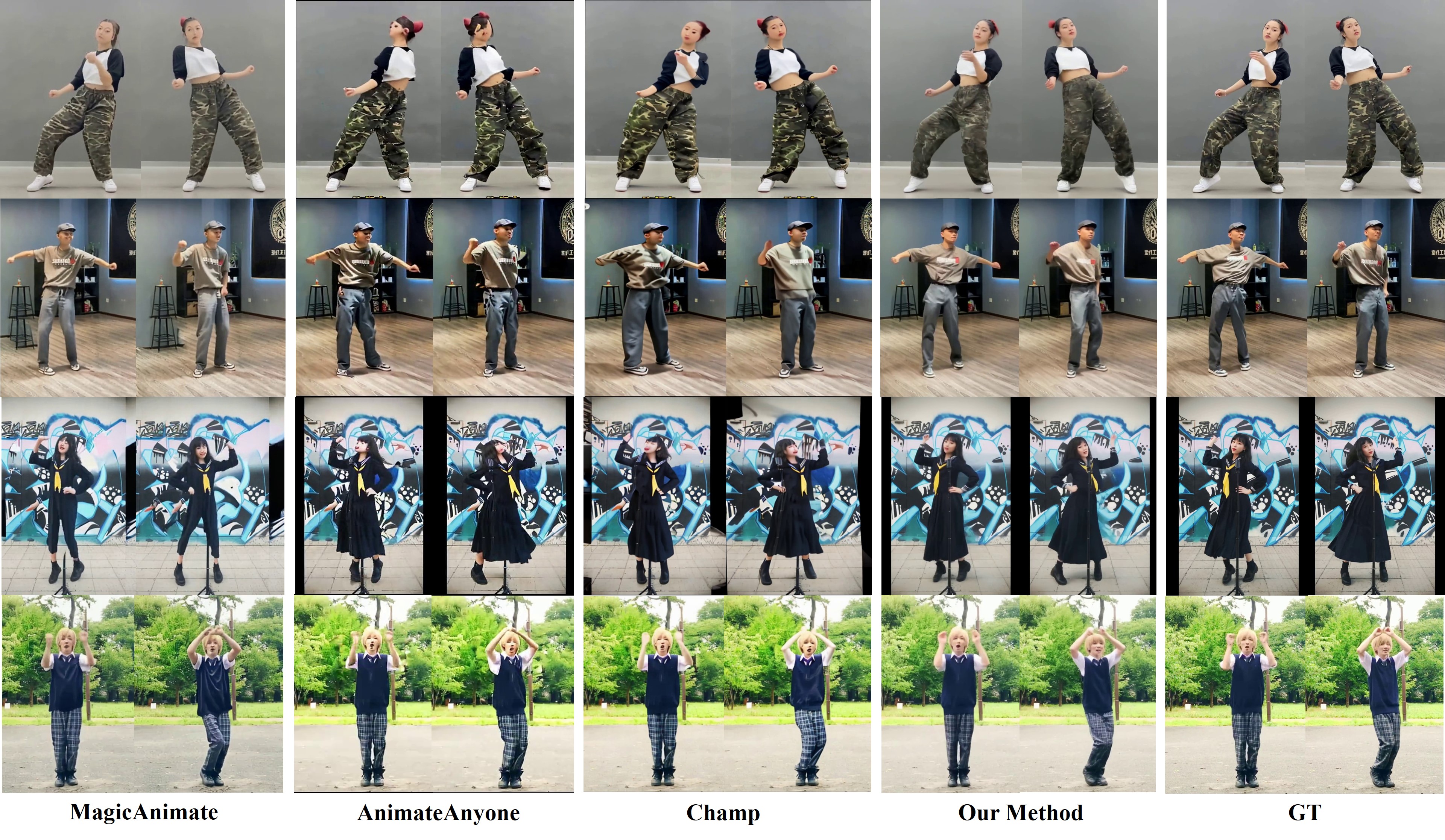}
    \vspace{-7mm}
    \caption{Qualitative comparison on monocular video.}
    \label{fig:video_comp}
\end{figure*}

\begin{table}[t]
\small
\centering
\begin{tabular}{c|cccc}
\hline
Method & PSNR $\uparrow$ & SSIM $\uparrow$ &LPIPS $\downarrow$ & FVD $\downarrow$ \\
\hline
Disco  & 20.07 & 0.661 & 0.285 & 585.3  \\
MagicAnimate  & 21.08 & 0.717 & 0.256 & 550.7  \\
AnimateAnyone  & 22.18 & 0.789 & 0.195 & 479.5  \\
Champ & 22.88 & 0.824 & 0.171 & 359.3  \\ \hline
AnimateAnyone*  & 24.27 & 0.862 & 0.148 & 341.0  \\
Champ* & 24.93 & 0.870 & 0.139 & 307.6 \\ \hline
Ours (small) & 25.45 & 0.877 & 0.121 & 276.1 \\
Ours (image) & 24.15 & 0.858 & 0.147 & 491.2 \\
Ours & \textbf{26.12} & \textbf{0.888} & \textbf{0.116} & \textbf{237.4} \\
\hline
\end{tabular}
\caption{Quantitative comparison on monocular video. * indicates that we fine-tuned the model on our dataset.  Ours (small) is the small version of our 4D transformer model with fewer layers. Ours (image) indicates that our model only has image transformer blocks.}
\label{tab:video_comp}
\vspace{-6mm}
\end{table}

\begin{table*}[t]
   \centering
   \begin{tabular*}{.95\linewidth}{c|ccc|ccc|ccc|ccc}
  \hline
  \multirow{2}*{Method}  & \multicolumn{3}{c|}{\makecell[c]{PSNR $\uparrow$ \\ }} & \multicolumn{3}{c|}{\makecell[c]{SSIM $\uparrow$ \\ }} & \multicolumn{3}{c|}{\makecell[c]{LPIPS $\downarrow$ \\ }} & \multicolumn{3}{c}{\makecell[c]{FVD $\downarrow$ \\ }} \\
    & MV & 3D & 4D &   MV & 3D & 4D& MV & 3D & 4D & MV & 3D & 4D \\ \hline
       \makecell[c]{Disco}  & 18.86&17.13&19.98 & 0.796&0.882&0.872 & 0.293&0.209&0.169 & 646.2&451.6&559.7 \\ 
       \makecell[c]{MagicAnimate}  & 19.30&19.28&21.74 & 0.845&0.906&0.920 & 0.232&0.159&0.135 & 517.3&356.4&418.4\\ 
       \makecell[c]{AnimateAnyone} & 19.87&20.53&22.01 & 0.858&0.922&0.912 & 0.216&0.111&0.131 & 472.8&285.4&410.3 \\ 
       \makecell[c]{Champ}  & 20.15&21.11&23.35 & 0.886&0.927&0.922 & 0.203&0.106&0.110 & 442.4&204.3&347.6  \\ \hline
       \makecell[c]{AnimateAnyone*} & 20.78 & 21.92 & 24.06 & 0.904 & 0.937 & 0.931 & 0.193 & 0.068 & 0.089 & 413.0 & 194.9 & 301.2 \\
       \makecell[c]{Champ*}  & 20.90 & 22.18 & 24.34 & 0.905 & 0.940 &  0.935 & 0.185 & 0.060 & 0.083 & 397.3 & 186.3 & 286.1 \\ \hline
       \makecell[c]{Ours (small)}  & 22.02 & 22.94 & 24.56 & 0.911 & 0.952 & 0.935 & 0.166 & 0.053 & 0.069 & 344.3 & 129.7 & 256.5 \\ 
       \makecell[c]{Ours (image)}  & 20.66 & 21.41 & 24.15 & 0.899 & 0.931 & 0.928 & 0.197 & 0.072 & 0.088 & 563.1 & 319.0 & 435.4 \\ 
       \makecell[c]{Ours (image+temporal)}  & 21.16 & 22.58 & 24.74 & 0.907 & 0.941 & 0.936 & 0.190 & 0.058 & 0.085 & 374.2 & 165.2 & 274.1 \\ 
       \makecell[c]{Ours} & \textbf{22.40} & \textbf{23.37} & \textbf{25.02} & \textbf{0.920} & \textbf{0.962} & \textbf{0.947} & \textbf{0.159} & \textbf{0.045} & \textbf{0.062} & \textbf{296.53} & \textbf{110.0} & \textbf{234.8} \\ 
       \hline
  \end{tabular*}
  \vspace{5pt}
   \caption{Quantitative comparison on multi-view (MV), 3D static (3D), and 360-degree (4D) videos. * indicates that we fine-tuned the model on our dataset. Ours (small) is the small version of our 4D transformer model with fewer layers. Ours (image) indicates that our model only has image transformer blocks. Ours (image+temporal) indicates our model without the view transformer blocks.}
   
\vspace{-6mm}
   \label{tab:dview_comp}
\end{table*}

\subsection{Implementation Details}
\subsubsection{Network Architecture}
Our method is based on the latent diffusion model that utilizes the VAE from Stable Diffusion XL. Our 4D diffusion transformer contains 10 4D transformer blocks, totaling 30 transformer layers.  Our CNNs-based encoder including $\mathbf{E}_{p}, \mathbf{E}_{id}$ compresses the input by a factor of 8, with each token having 1280 channels, and the text embedding and image identity embedding are also mapped to 1280 channels. The latent encoder $\mathbf{E}_\mathbf{z}$ further compresses the input by a factor of 2. During training, the video data is resized to 768x768 resolution, and for each GPU, the input consists of 24 frames. When training on image data, we use a batch size of 24, while for monocular video training, we use a batch size of 1 with a video length of 24. For multi-view video, 3D video, and 360-degree video training, we use a batch size of 1 with a video length of 6 and 4 views. 
\subsubsection{Training Details}
We employed 24 A100 GPUs, and the total training time was 14 days.
We train our model with a learning rate 1e-5, first on the image dataset for 50k iterations, then on the full dataset for 150k iterations. We use ColossalAI's HybridAdam optimizer with Zero2 strategy and zero weight decay to optimize parameters. We also clip the gradient in the range of [-1, 1].

\subsection{Comparisons}
\subsubsection{Comparisons on Monocular Video.}
For monocular video, we randomly select 200 videos from the Human4DiT-Video dataset as our test set for comparison, with the first frame of each video serving as the reference image. During inference, our method employs a temporal window of 24 frames, with an overlap of 6 frames between consecutive windows. For Disco~\cite{wang2023disco}, MagicAnimate~\cite{xu2023magicanimate}, and Champ~\cite{zhu2024champ}, we use their official open-source code. 
As for AnimateAnyone~\cite{hu2023animate}, we employ the opensource implementation from MooreThreads~\footnote{https://github.com/MooreThreads/Moore-AnimateAnyone}
. To ensure a more fair comparison, we fine-tuned AnimateAnyone and Champ on our Human4DiT dataset for 10k iterations. Results from these fine-tuned models are denoted with an asterisk (*) in our evaluations.
Quantitative comparisons are presented in Tab.~\ref{tab:video_comp}, where our method demonstrates a clear numerical advantage, significantly outperforming other approaches. This highlights the superiority of our 4D diffusion transformer over CNNs-based methods. We have also conducted qualitative comparisons, with results shown in Fig.~\ref{fig:video_comp}. Our method generates more natural dynamic effects with fewer deformation and jitter artifacts compared to other methods, indicating the 4D transformer's stronger capability in establishing spatial-temporal consistency than U-Net-based approaches. Please refer to our submitted project webpage for dynamic video effects.

\begin{figure}
    \centering
    \includegraphics[width=\linewidth]{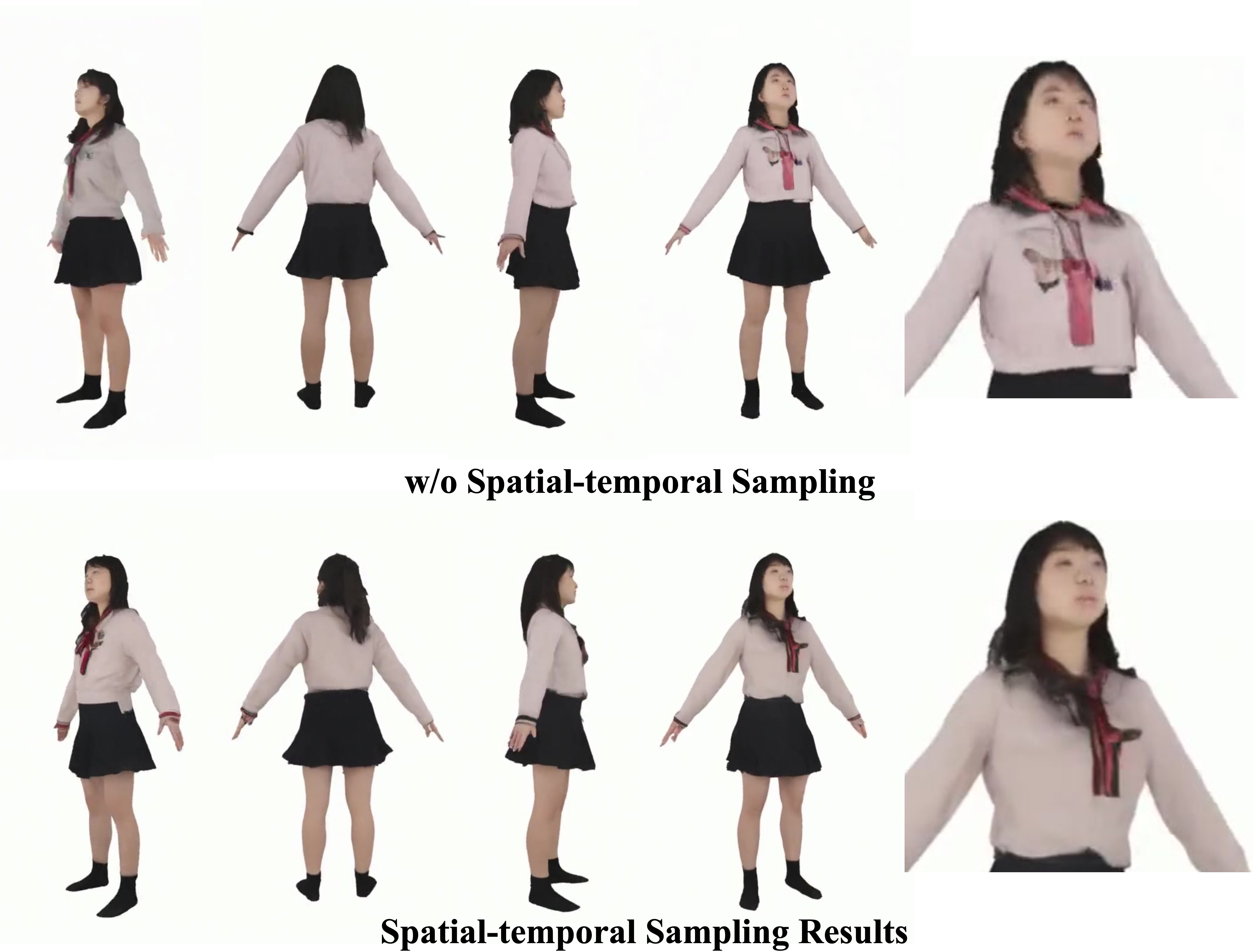}
    \caption{Ablation study of spatial-temporal sampling.}
    \label{fig:ablation_st}
\end{figure}

\subsubsection{Comparisons on Multi-view Video.}
For multi-view settings, we select 25 multi-view groups from the AIST and DNA-rendering datasets respectively as our test set for comparison, with each group containing videos captured from 4 different viewpoints. We use the first frame of the frontal view as the reference image. During inference, our method employs the spatial-temporal sampling strategy, with a spatial window size of 4 views and a temporal window of 6 frames for spatial sampling. For temporal sampling, the window size is 24 frames. The blending weights for spatial and temporal $\lambda_1, \lambda_2$ are set to 0.5 and 0.5, respectively. For Disco, MagicAnimate, Champ, and AnimateAnyone, we treat each view's video as a monocular video and perform inference separately. Quantitative comparisons are presented in Tab.~\ref{tab:dview_comp}, where we masked out the backgrounds for metric computation since inferring other views' backgrounds from a single view is an ill-posed problem. Our method outperforms others, demonstrating the 4D diffusion transformer's superior ability to establish stronger cross-view correlations compared to U-Net-based approaches. We have also conducted qualitative comparisons, with results shown in Fig.~\ref{fig:mv_comp}. Our method generates spatio-temporally consistent multi-view videos without exhibiting multi-face artifacts. Please refer to the project webpage for dynamic video effects.

\subsubsection{Comparisons on 3D Static Video.}
For 3D setting, we select 100 3D scans from THUman2.0 and render them into 3D static videos as our test set for comparison. We use the frontal view as the reference image. During inference, our method employs the same spatial-temporal sampling strategy as the multi-view setting. Quantitative comparisons are presented in Tab.~\ref{tab:dview_comp}. We mask out the backgrounds for a fair comparison since other methods tend to generate noisy backgrounds. Our method outperforms others, demonstrating the 4D diffusion transformer's ability to learn physical 3D viewpoint changes. We have also conducted qualitative comparisons, with results shown in Fig.~\ref{fig:3d_comp}. Our method generates spatially consistent 360-degree videos compared to other approaches. Please refer to our submitted project webpage for 3D static video effects.

\subsubsection{Comparisons on 360-degree Video.}
For 360-degree video evaluation, we select 10 4D scans from the Human4DiT dataset, each performing 5 different motions, and render them with different camera trajectories to create 50 test videos. We use the frontal view as the reference image. During inference, our method employs the same spatial-temporal sampling strategy as the multi-view setting. Quantitative comparisons are presented in Tab.~\ref{tab:dview_comp}, where we mask out the backgrounds for the evaluation. The results demonstrate our 4D diffusion transformer's ability to handle both viewpoint and human motion changes in a 4D dynamic scenario. We have also conducted qualitative comparisons, with results shown in Fig.~\ref{fig:4d_comp}. Our method generates spatio-temporally consistent 360-degree dynamic videos compared to other approaches. Please refer to our submitted project webpage for 360-degree video effects.

\subsubsection{Comparisons on model size.} In addition to comparing video generation quality, we conducted a comprehensive analysis of model sizes across various approaches, as illustrated in Tab~\ref{tab:params_comp}. We also train a smaller version of Human4DiT "Ours (small)" with 6 4D transformer blocks. Our Human4DiT model in its small setting exhibits comparable parameter counts to AnimateAnyone and Champ. However, due to the higher computational complexity of attention operations in the DiT architecture compared to convolutional operations, our model does not demonstrate an advantage in terms of inference speed. Nonetheless, as presented by our previous experimental comparisons, our model achieves superior performance within fewer parameters, thus validating the efficiency of our proposed Human4DiT architecture.

\begin{figure}
    \centering
    \includegraphics[width=\linewidth]{figs/temporalablation.jpg}
    \vspace{-6mm}
    \caption{Ablation study of temporal transformer. When using only an image transformer without a temporal transformer, temporal consistency could not be guaranteed, resulting in artifacts such as discontinuous arm generation (see top images). After introducing the temporal transformer, our method can produce continuous and natural human motion (see bottom images).}
    \label{fig:ablation_temporal}
\end{figure}

\begin{figure}
    \centering
    \includegraphics[width=\linewidth]{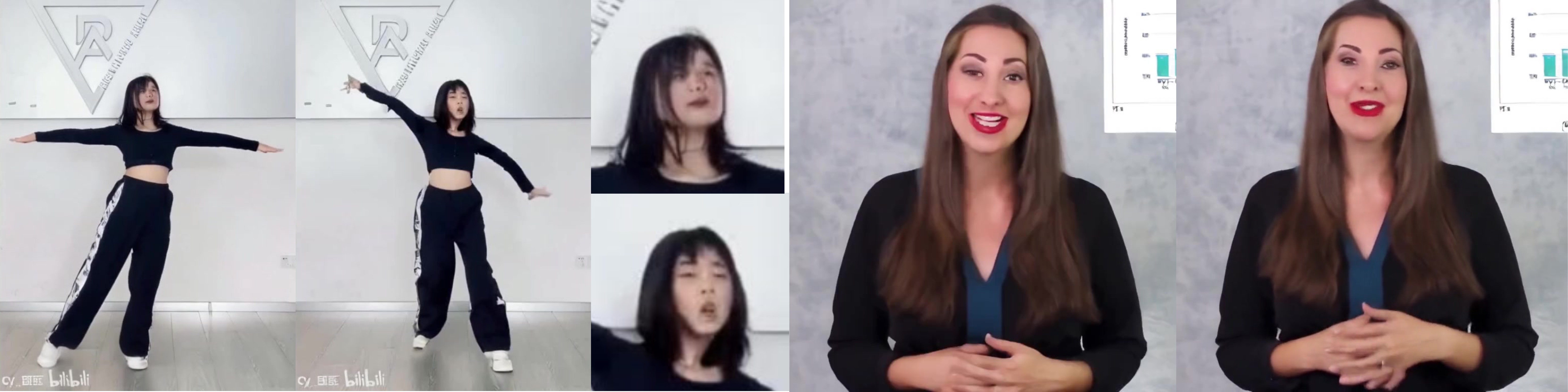}
    \vspace{-4mm}
    \caption{Failure cases. Our method produces distortions and artifacts in cases where the facial region is small (see left images). Additionally, our approach struggles to generate natural finger structures (see right images).}
    \label{fig:failure}
    \vspace{-4mm}
\end{figure}

\begin{table}[t]
\small
\centering
{\color{black}
\begin{tabular}{c|cccc}

\hline
Method & Parameters & T-size & Resolution & FPS \\
\hline
AnimateAnyone  & 2.07B & 24 & 512x512 & 0.769  \\
Champ & 2.04B & 24 & 768x768 & 0.657  \\
Ours (small)  & 1.83B & 24 & 768x768 & 0.605  \\
Ours & 3.10B & 24 & 768x768 & 0.443 \\
\hline
\end{tabular}
\caption{Performance comparisons of different video models. T-size is the temporal window size. FPS is the average number of generated frames every second with 25 inference steps. }
 \label{tab:params_comp}
 
\vspace{-6mm}
}
\end{table}

\subsection{Ablation Study}
{\color{black}
\subsubsection{Temporal Transformer.} First, we validate our temporal transformer blocks and conduct an ablation study by removing the temporal transformers. The quantitative and qualitative results "Ours (image)" are presented in Tab.~\ref{tab:dview_comp} and Fig.~\ref{fig:ablation_temporal}. 
Observations indicate that the incorporation of temporal transformers significantly enhances generation performance. Notably, even in the absence of these components, our method demonstrates comparable performance to CNN-based approaches.
\subsubsection{View Transformer}
To validate the effectiveness of introducing the fourth view dimension and the view transformer in our 4D transformer, we conduct ablation studies for multi-view video, 3D static video, and 360-degree video generation tasks. Specifically, we disable the view transformer and rely solely on the temporal transformer to generate the corresponding videos. Quantitative results of "Ours (image+temporal)" are presented in Tab.~\ref{tab:dview_comp}.
It can be observed that the introduction of the view transformer leads to improvements in view-related generation tasks, demonstrating the efficacy of the view transformer.
}

{\color{black}
\subsubsection{Efficient Spatial-temporal Sampling}
To validate the efficacy of our spatial-temporal sampling, we conducted a qualitative comparison, as illustrated in Fig~\ref{fig:ablation_st}. When only temporal window sampling is employed, the generated 360-degree video exhibits noticeable misalignments (The tie on the person's chest has become a flat, texture surface). In contrast, our spatial-temporal sampling method mitigates these inconsistencies, achieving global coherence throughout the generated sequence.
 The results demonstrate that our method effectively addresses the challenges of long-range dependencies in 360-degree human video generation, contributing to enhanced visual quality and coherence.

\subsection{Failure Cases}\label{sec:fail} As illustrated in Fig.~\ref{fig:failure}, our work primarily addresses full-body, 360-degree human video generation. However, the face and hands of human, typically occupying less than 100x100 pixels within our 768x768 generated video frames, are leading to artifacts due to their relatively small scale. Furthermore, inaccuracies in SMPL estimation also affect the overall generation quality. We believe that future research incorporating specialized high-resolution modules for facial and hand generation could potentially address these issues.

Additionally, our current approach demonstrates limitations in effectively modeling background dynamics. This problem arises from the nature of our dataset, which predominantly features static backgrounds. Moreover, our 3D/4D datasets are rendered against a uniform white background. Consequently, the model has not learned to accurately represent background transformations across varying viewpoints.
\section{Discussion}

\noindent\textbf{Conclusion.} 
We have presented a novel method for human video generation, which takes as input only a single image and produces spatio-temporally coherent video of dynamic human motions under 360-degree viewpoints. Our approach employs an efficient 4D transformer architecture to model the correlations across multiple domains, including view, time and poses. Combining with UNets for accurate condition injection, our model can be trained on a multi-dimensional dataset spanning images, videos, multi-view data, and 4D scans. After training, our method can synthesize 360-degree realistic, coherent human motion videos, and we believe our contributions will inspire future work towards 4D content generation.

\noindent \textbf{Limitations and Future Work.} 
Instead of generating an explicit 4D models, our method directly synthesizes 2D videos from the given viewpoints, and the 4D scene structure are implicitly encoded through the attention mechanism. The absence of an explicit 4D representation results in some artifacts when rendering 360-degree videos. 
Furthermore, our current implementation cannot generate tiny structures coherently as mentioned in Sec.~\ref{sec:fail}. Future works may incorporate a body part-aware generator to resolve this limitation. Exploring other condition injection techniques is also significant to enhance controllability. Potential directions include improved camera representations, Plücker embeddings, and direct utilization of 3D SMPL as tokens. We also believe collecting a more diverse dataset is significant to fully leverage the capabilities of DiT, and address video generation challenges in out-of-distribution (OOD) scenarios, such as anime or stylized content. 

\section{Ethics Statement}
While our method for human video generation could benefit a lot of applications for entertainment, education, and accessibility, we recognize the risks of misuse for creating deepfakes, privacy violations, and perpetuating societal biases. We are committed to responsible development and promoting transparency in our methods. By discussing these ethical considerations, we aim to foster the positive applications of this technology while minimizing potential harm.

\noindent\textbf{Acknowledgement}
The work is supported by the National Science Foundation of China under Grant Number 62125107.

\bibliographystyle{ACM-Reference-Format}
\bibliography{sample-bibliography}


\begin{thebibliography}{52}


\ifx \showCODEN    \undefined \def \showCODEN     #1{\unskip}     \fi
\ifx \showDOI      \undefined \def \showDOI       #1{#1}\fi
\ifx \showISBNx    \undefined \def \showISBNx     #1{\unskip}     \fi
\ifx \showISBNxiii \undefined \def \showISBNxiii  #1{\unskip}     \fi
\ifx \showISSN     \undefined \def \showISSN      #1{\unskip}     \fi
\ifx \showLCCN     \undefined \def \showLCCN      #1{\unskip}     \fi
\ifx \shownote     \undefined \def \shownote      #1{#1}          \fi
\ifx \showarticletitle \undefined \def \showarticletitle #1{#1}   \fi
\ifx \showURL      \undefined \def \showURL       {\relax}        \fi
\providecommand\bibfield[2]{#2}
\providecommand\bibinfo[2]{#2}
\providecommand\natexlab[1]{#1}
\providecommand\showeprint[2][]{arXiv:#2}

\bibitem[Bao et~al\mbox{.}(2023)]%
        {bao2023all}
\bibfield{author}{\bibinfo{person}{Fan Bao}, \bibinfo{person}{Shen Nie}, \bibinfo{person}{Kaiwen Xue}, \bibinfo{person}{Yue Cao}, \bibinfo{person}{Chongxuan Li}, \bibinfo{person}{Hang Su}, {and} \bibinfo{person}{Jun Zhu}.} \bibinfo{year}{2023}\natexlab{}.
\newblock \showarticletitle{All are worth words: A vit backbone for diffusion models}. In \bibinfo{booktitle}{\emph{Proceedings of the IEEE/CVF Conference on Computer Vision and Pattern Recognition}}. \bibinfo{pages}{22669--22679}.
\newblock


\bibitem[Bhunia et~al\mbox{.}(2023)]%
        {bhunia2023person}
\bibfield{author}{\bibinfo{person}{Ankan~Kumar Bhunia}, \bibinfo{person}{Salman Khan}, \bibinfo{person}{Hisham Cholakkal}, \bibinfo{person}{Rao~Muhammad Anwer}, \bibinfo{person}{Jorma Laaksonen}, \bibinfo{person}{Mubarak Shah}, {and} \bibinfo{person}{Fahad~Shahbaz Khan}.} \bibinfo{year}{2023}\natexlab{}.
\newblock \showarticletitle{Person image synthesis via denoising diffusion model}. In \bibinfo{booktitle}{\emph{Proceedings of the IEEE/CVF Conference on Computer Vision and Pattern Recognition}}. \bibinfo{pages}{5968--5976}.
\newblock


\bibitem[Black et~al\mbox{.}(2023)]%
        {black2023bedlam}
\bibfield{author}{\bibinfo{person}{Michael~J Black}, \bibinfo{person}{Priyanka Patel}, \bibinfo{person}{Joachim Tesch}, {and} \bibinfo{person}{Jinlong Yang}.} \bibinfo{year}{2023}\natexlab{}.
\newblock \showarticletitle{Bedlam: A synthetic dataset of bodies exhibiting detailed lifelike animated motion}. In \bibinfo{booktitle}{\emph{Proceedings of the IEEE/CVF Conference on Computer Vision and Pattern Recognition}}. \bibinfo{pages}{8726--8737}.
\newblock


\bibitem[Cheng et~al\mbox{.}(2023)]%
        {cheng2023dna}
\bibfield{author}{\bibinfo{person}{Wei Cheng}, \bibinfo{person}{Ruixiang Chen}, \bibinfo{person}{Siming Fan}, \bibinfo{person}{Wanqi Yin}, \bibinfo{person}{Keyu Chen}, \bibinfo{person}{Zhongang Cai}, \bibinfo{person}{Jingbo Wang}, \bibinfo{person}{Yang Gao}, \bibinfo{person}{Zhengming Yu}, \bibinfo{person}{Zhengyu Lin}, {et~al\mbox{.}}} \bibinfo{year}{2023}\natexlab{}.
\newblock \showarticletitle{Dna-rendering: A diverse neural actor repository for high-fidelity human-centric rendering}. In \bibinfo{booktitle}{\emph{Proceedings of the IEEE/CVF International Conference on Computer Vision}}. \bibinfo{pages}{19982--19993}.
\newblock


\bibitem[Goel et~al\mbox{.}(2023)]%
        {goel2023humansIn4d}
\bibfield{author}{\bibinfo{person}{Shubham Goel}, \bibinfo{person}{Georgios Pavlakos}, \bibinfo{person}{Jathushan Rajasegaran}, \bibinfo{person}{Angjoo Kanazawa*}, {and} \bibinfo{person}{Jitendra Malik*}.} \bibinfo{year}{2023}\natexlab{}.
\newblock \showarticletitle{Humans in 4{D}: Reconstructing and Tracking Humans with Transformers}. In \bibinfo{booktitle}{\emph{International Conference on Computer Vision (ICCV)}}.
\newblock


\bibitem[Goodfellow et~al\mbox{.}(2014)]%
        {goodfellow2014generative}
\bibfield{author}{\bibinfo{person}{Ian Goodfellow}, \bibinfo{person}{Jean Pouget-Abadie}, \bibinfo{person}{Mehdi Mirza}, \bibinfo{person}{Bing Xu}, \bibinfo{person}{David Warde-Farley}, \bibinfo{person}{Sherjil Ozair}, \bibinfo{person}{Aaron Courville}, {and} \bibinfo{person}{Yoshua Bengio}.} \bibinfo{year}{2014}\natexlab{}.
\newblock \showarticletitle{Generative adversarial nets}.
\newblock \bibinfo{journal}{\emph{Advances in neural information processing systems}}  \bibinfo{volume}{27} (\bibinfo{year}{2014}).
\newblock


\bibitem[He et~al\mbox{.}(2024)]%
        {he2024cameractrl}
\bibfield{author}{\bibinfo{person}{Hao He}, \bibinfo{person}{Yinghao Xu}, \bibinfo{person}{Yuwei Guo}, \bibinfo{person}{Gordon Wetzstein}, \bibinfo{person}{Bo Dai}, \bibinfo{person}{Hongsheng Li}, {and} \bibinfo{person}{Ceyuan Yang}.} \bibinfo{year}{2024}\natexlab{}.
\newblock \bibinfo{title}{CameraCtrl: Enabling Camera Control for Text-to-Video Generation}.
\newblock
\newblock
\showeprint[arxiv]{2404.02101}~[cs.CV]


\bibitem[Hu et~al\mbox{.}(2023)]%
        {hu2023animate}
\bibfield{author}{\bibinfo{person}{Li Hu}, \bibinfo{person}{Xin Gao}, \bibinfo{person}{Peng Zhang}, \bibinfo{person}{Ke Sun}, \bibinfo{person}{Bang Zhang}, {and} \bibinfo{person}{Liefeng Bo}.} \bibinfo{year}{2023}\natexlab{}.
\newblock \showarticletitle{Animate anyone: Consistent and controllable image-to-video synthesis for character animation}.
\newblock \bibinfo{journal}{\emph{arXiv preprint arXiv:2311.17117}} (\bibinfo{year}{2023}).
\newblock


\bibitem[Jafarian and Park(2021)]%
        {jafarian2021learning}
\bibfield{author}{\bibinfo{person}{Yasamin Jafarian} {and} \bibinfo{person}{Hyun~Soo Park}.} \bibinfo{year}{2021}\natexlab{}.
\newblock \showarticletitle{Learning high fidelity depths of dressed humans by watching social media dance videos}. In \bibinfo{booktitle}{\emph{Proceedings of the IEEE/CVF Conference on Computer Vision and Pattern Recognition}}. \bibinfo{pages}{12753--12762}.
\newblock


\bibitem[Ju et~al\mbox{.}(2023)]%
        {ju2023human}
\bibfield{author}{\bibinfo{person}{Xuan Ju}, \bibinfo{person}{Ailing Zeng}, \bibinfo{person}{Jianan Wang}, \bibinfo{person}{Qiang Xu}, {and} \bibinfo{person}{Lei Zhang}.} \bibinfo{year}{2023}\natexlab{}.
\newblock \showarticletitle{Human-art: A versatile human-centric dataset bridging natural and artificial scenes}. In \bibinfo{booktitle}{\emph{Proceedings of the IEEE/CVF Conference on Computer Vision and Pattern Recognition}}. \bibinfo{pages}{618--629}.
\newblock


\bibitem[Karras et~al\mbox{.}(2023)]%
        {karras2023dreampose}
\bibfield{author}{\bibinfo{person}{Johanna Karras}, \bibinfo{person}{Aleksander Holynski}, \bibinfo{person}{Ting-Chun Wang}, {and} \bibinfo{person}{Ira Kemelmacher-Shlizerman}.} \bibinfo{year}{2023}\natexlab{}.
\newblock \showarticletitle{Dreampose: Fashion video synthesis with stable diffusion}. In \bibinfo{booktitle}{\emph{Proceedings of the IEEE/CVF International Conference on Computer Vision}}. \bibinfo{pages}{22680--22690}.
\newblock


\bibitem[Lee et~al\mbox{.}(2023)]%
        {lee2023same}
\bibfield{author}{\bibinfo{person}{Sunmin Lee}, \bibinfo{person}{Taeho Kang}, \bibinfo{person}{Jungnam Park}, \bibinfo{person}{Jehee Lee}, {and} \bibinfo{person}{Jungdam Won}.} \bibinfo{year}{2023}\natexlab{}.
\newblock \showarticletitle{SAME: Skeleton-Agnostic Motion Embedding for Character Animation}. In \bibinfo{booktitle}{\emph{SIGGRAPH Asia 2023 Conference Papers}}. \bibinfo{pages}{1--11}.
\newblock


\bibitem[Lin et~al\mbox{.}(2024)]%
        {lin2024motion}
\bibfield{author}{\bibinfo{person}{Jing Lin}, \bibinfo{person}{Ailing Zeng}, \bibinfo{person}{Shunlin Lu}, \bibinfo{person}{Yuanhao Cai}, \bibinfo{person}{Ruimao Zhang}, \bibinfo{person}{Haoqian Wang}, {and} \bibinfo{person}{Lei Zhang}.} \bibinfo{year}{2024}\natexlab{}.
\newblock \showarticletitle{Motion-x: A large-scale 3d expressive whole-body human motion dataset}.
\newblock \bibinfo{journal}{\emph{Advances in Neural Information Processing Systems}}  \bibinfo{volume}{36} (\bibinfo{year}{2024}).
\newblock


\bibitem[Liu et~al\mbox{.}(2023b)]%
        {liu2023zero}
\bibfield{author}{\bibinfo{person}{Ruoshi Liu}, \bibinfo{person}{Rundi Wu}, \bibinfo{person}{Basile Van~Hoorick}, \bibinfo{person}{Pavel Tokmakov}, \bibinfo{person}{Sergey Zakharov}, {and} \bibinfo{person}{Carl Vondrick}.} \bibinfo{year}{2023}\natexlab{b}.
\newblock \showarticletitle{Zero-1-to-3: Zero-shot one image to 3d object}. In \bibinfo{booktitle}{\emph{Proceedings of the IEEE/CVF International Conference on Computer Vision}}. \bibinfo{pages}{9298--9309}.
\newblock


\bibitem[Liu et~al\mbox{.}(2023a)]%
        {liu2023syncdreamer}
\bibfield{author}{\bibinfo{person}{Yuan Liu}, \bibinfo{person}{Cheng Lin}, \bibinfo{person}{Zijiao Zeng}, \bibinfo{person}{Xiaoxiao Long}, \bibinfo{person}{Lingjie Liu}, \bibinfo{person}{Taku Komura}, {and} \bibinfo{person}{Wenping Wang}.} \bibinfo{year}{2023}\natexlab{a}.
\newblock \showarticletitle{Syncdreamer: Generating multiview-consistent images from a single-view image}.
\newblock \bibinfo{journal}{\emph{arXiv preprint arXiv:2309.03453}} (\bibinfo{year}{2023}).
\newblock


\bibitem[Long et~al\mbox{.}(2023)]%
        {long2023wonder3d}
\bibfield{author}{\bibinfo{person}{Xiaoxiao Long}, \bibinfo{person}{Yuan-Chen Guo}, \bibinfo{person}{Cheng Lin}, \bibinfo{person}{Yuan Liu}, \bibinfo{person}{Zhiyang Dou}, \bibinfo{person}{Lingjie Liu}, \bibinfo{person}{Yuexin Ma}, \bibinfo{person}{Song-Hai Zhang}, \bibinfo{person}{Marc Habermann}, \bibinfo{person}{Christian Theobalt}, {et~al\mbox{.}}} \bibinfo{year}{2023}\natexlab{}.
\newblock \showarticletitle{Wonder3d: Single image to 3d using cross-domain diffusion}.
\newblock \bibinfo{journal}{\emph{arXiv preprint arXiv:2310.15008}} (\bibinfo{year}{2023}).
\newblock


\bibitem[Loper et~al\mbox{.}(2023)]%
        {loper2023smpl}
\bibfield{author}{\bibinfo{person}{Matthew Loper}, \bibinfo{person}{Naureen Mahmood}, \bibinfo{person}{Javier Romero}, \bibinfo{person}{Gerard Pons-Moll}, {and} \bibinfo{person}{Michael~J Black}.} \bibinfo{year}{2023}\natexlab{}.
\newblock \showarticletitle{SMPL: A skinned multi-person linear model}.
\newblock In \bibinfo{booktitle}{\emph{Seminal Graphics Papers: Pushing the Boundaries, Volume 2}}. \bibinfo{pages}{851--866}.
\newblock


\bibitem[Lu et~al\mbox{.}(2023)]%
        {lu2023vdt}
\bibfield{author}{\bibinfo{person}{Haoyu Lu}, \bibinfo{person}{Guoxing Yang}, \bibinfo{person}{Nanyi Fei}, \bibinfo{person}{Yuqi Huo}, \bibinfo{person}{Zhiwu Lu}, \bibinfo{person}{Ping Luo}, {and} \bibinfo{person}{Mingyu Ding}.} \bibinfo{year}{2023}\natexlab{}.
\newblock \showarticletitle{Vdt: General-purpose video diffusion transformers via mask modeling}. In \bibinfo{booktitle}{\emph{The Twelfth International Conference on Learning Representations}}.
\newblock


\bibitem[Ma et~al\mbox{.}(2024)]%
        {ma2024latte}
\bibfield{author}{\bibinfo{person}{Xin Ma}, \bibinfo{person}{Yaohui Wang}, \bibinfo{person}{Gengyun Jia}, \bibinfo{person}{Xinyuan Chen}, \bibinfo{person}{Ziwei Liu}, \bibinfo{person}{Yuan-Fang Li}, \bibinfo{person}{Cunjian Chen}, {and} \bibinfo{person}{Yu Qiao}.} \bibinfo{year}{2024}\natexlab{}.
\newblock \showarticletitle{Latte: Latent diffusion transformer for video generation}.
\newblock \bibinfo{journal}{\emph{arXiv preprint arXiv:2401.03048}} (\bibinfo{year}{2024}).
\newblock


\bibitem[Mirza and Osindero(2014)]%
        {mirza2014conditional}
\bibfield{author}{\bibinfo{person}{Mehdi Mirza} {and} \bibinfo{person}{Simon Osindero}.} \bibinfo{year}{2014}\natexlab{}.
\newblock \showarticletitle{Conditional generative adversarial nets}.
\newblock \bibinfo{journal}{\emph{arXiv preprint arXiv:1411.1784}} (\bibinfo{year}{2014}).
\newblock


\bibitem[OpenAI(2024)]%
        {sora2024}
\bibfield{author}{\bibinfo{person}{OpenAI}.} \bibinfo{year}{2024}\natexlab{}.
\newblock \bibinfo{title}{Video generation models as world simulators}.
\newblock \bibinfo{howpublished}{\url{https://openai.com/index/video-generation-models-as-world-simulators/}}.
\newblock
\newblock
\shownote{Accessed: 2024-05-19}.


\bibitem[Peebles and Xie(2023)]%
        {peebles2023scalable}
\bibfield{author}{\bibinfo{person}{William Peebles} {and} \bibinfo{person}{Saining Xie}.} \bibinfo{year}{2023}\natexlab{}.
\newblock \showarticletitle{Scalable diffusion models with transformers}. In \bibinfo{booktitle}{\emph{Proceedings of the IEEE/CVF International Conference on Computer Vision}}. \bibinfo{pages}{4195--4205}.
\newblock


\bibitem[Radford et~al\mbox{.}(2018)]%
        {radford2018improving}
\bibfield{author}{\bibinfo{person}{Alec Radford}, \bibinfo{person}{Karthik Narasimhan}, \bibinfo{person}{Tim Salimans}, \bibinfo{person}{Ilya Sutskever}, {et~al\mbox{.}}} \bibinfo{year}{2018}\natexlab{}.
\newblock \showarticletitle{Improving language understanding by generative pre-training}.
\newblock  (\bibinfo{year}{2018}).
\newblock


\bibitem[Radford et~al\mbox{.}(2019)]%
        {radford2019language}
\bibfield{author}{\bibinfo{person}{Alec Radford}, \bibinfo{person}{Jeffrey Wu}, \bibinfo{person}{Rewon Child}, \bibinfo{person}{David Luan}, \bibinfo{person}{Dario Amodei}, \bibinfo{person}{Ilya Sutskever}, {et~al\mbox{.}}} \bibinfo{year}{2019}\natexlab{}.
\newblock \showarticletitle{Language models are unsupervised multitask learners}.
\newblock \bibinfo{journal}{\emph{OpenAI blog}} \bibinfo{volume}{1}, \bibinfo{number}{8} (\bibinfo{year}{2019}), \bibinfo{pages}{9}.
\newblock


\bibitem[Rombach et~al\mbox{.}(2022)]%
        {rombach2022high}
\bibfield{author}{\bibinfo{person}{Robin Rombach}, \bibinfo{person}{Andreas Blattmann}, \bibinfo{person}{Dominik Lorenz}, \bibinfo{person}{Patrick Esser}, {and} \bibinfo{person}{Bj{\"o}rn Ommer}.} \bibinfo{year}{2022}\natexlab{}.
\newblock \showarticletitle{High-resolution image synthesis with latent diffusion models}. In \bibinfo{booktitle}{\emph{Proceedings of the IEEE/CVF conference on computer vision and pattern recognition}}. \bibinfo{pages}{10684--10695}.
\newblock


\bibitem[Ronneberger et~al\mbox{.}(2015)]%
        {ronneberger2015u}
\bibfield{author}{\bibinfo{person}{Olaf Ronneberger}, \bibinfo{person}{Philipp Fischer}, {and} \bibinfo{person}{Thomas Brox}.} \bibinfo{year}{2015}\natexlab{}.
\newblock \showarticletitle{U-net: Convolutional networks for biomedical image segmentation}. In \bibinfo{booktitle}{\emph{Medical image computing and computer-assisted intervention--MICCAI 2015: 18th international conference, Munich, Germany, October 5-9, 2015, proceedings, part III 18}}. Springer, \bibinfo{pages}{234--241}.
\newblock


\bibitem[Shi et~al\mbox{.}(2023a)]%
        {shi2023zero123++}
\bibfield{author}{\bibinfo{person}{Ruoxi Shi}, \bibinfo{person}{Hansheng Chen}, \bibinfo{person}{Zhuoyang Zhang}, \bibinfo{person}{Minghua Liu}, \bibinfo{person}{Chao Xu}, \bibinfo{person}{Xinyue Wei}, \bibinfo{person}{Linghao Chen}, \bibinfo{person}{Chong Zeng}, {and} \bibinfo{person}{Hao Su}.} \bibinfo{year}{2023}\natexlab{a}.
\newblock \showarticletitle{Zero123++: a single image to consistent multi-view diffusion base model}.
\newblock \bibinfo{journal}{\emph{arXiv preprint arXiv:2310.15110}} (\bibinfo{year}{2023}).
\newblock


\bibitem[Shi et~al\mbox{.}(2023b)]%
        {shi2023mvdream}
\bibfield{author}{\bibinfo{person}{Yichun Shi}, \bibinfo{person}{Peng Wang}, \bibinfo{person}{Jianglong Ye}, \bibinfo{person}{Mai Long}, \bibinfo{person}{Kejie Li}, {and} \bibinfo{person}{Xiao Yang}.} \bibinfo{year}{2023}\natexlab{b}.
\newblock \showarticletitle{Mvdream: Multi-view diffusion for 3d generation}.
\newblock \bibinfo{journal}{\emph{arXiv preprint arXiv:2308.16512}} (\bibinfo{year}{2023}).
\newblock


\bibitem[Siarohin et~al\mbox{.}(2019a)]%
        {siarohin2019appearance}
\bibfield{author}{\bibinfo{person}{Aliaksandr Siarohin}, \bibinfo{person}{St{\'e}phane Lathuili{\`e}re}, \bibinfo{person}{Enver Sangineto}, {and} \bibinfo{person}{Nicu Sebe}.} \bibinfo{year}{2019}\natexlab{a}.
\newblock \showarticletitle{Appearance and pose-conditioned human image generation using deformable gans}.
\newblock \bibinfo{journal}{\emph{IEEE transactions on pattern analysis and machine intelligence}} \bibinfo{volume}{43}, \bibinfo{number}{4} (\bibinfo{year}{2019}), \bibinfo{pages}{1156--1171}.
\newblock


\bibitem[Siarohin et~al\mbox{.}(2019b)]%
        {siarohin2019first}
\bibfield{author}{\bibinfo{person}{Aliaksandr Siarohin}, \bibinfo{person}{St{\'e}phane Lathuili{\`e}re}, \bibinfo{person}{Sergey Tulyakov}, \bibinfo{person}{Elisa Ricci}, {and} \bibinfo{person}{Nicu Sebe}.} \bibinfo{year}{2019}\natexlab{b}.
\newblock \showarticletitle{First order motion model for image animation}.
\newblock \bibinfo{journal}{\emph{Advances in neural information processing systems}}  \bibinfo{volume}{32} (\bibinfo{year}{2019}).
\newblock


\bibitem[Siarohin et~al\mbox{.}(2018)]%
        {siarohin2018deformable}
\bibfield{author}{\bibinfo{person}{Aliaksandr Siarohin}, \bibinfo{person}{Enver Sangineto}, \bibinfo{person}{St{\'e}phane Lathuiliere}, {and} \bibinfo{person}{Nicu Sebe}.} \bibinfo{year}{2018}\natexlab{}.
\newblock \showarticletitle{Deformable gans for pose-based human image generation}. In \bibinfo{booktitle}{\emph{Proceedings of the IEEE conference on computer vision and pattern recognition}}. \bibinfo{pages}{3408--3416}.
\newblock


\bibitem[Strudel et~al\mbox{.}(2021)]%
        {strudel2021segmenter}
\bibfield{author}{\bibinfo{person}{Robin Strudel}, \bibinfo{person}{Ricardo Garcia}, \bibinfo{person}{Ivan Laptev}, {and} \bibinfo{person}{Cordelia Schmid}.} \bibinfo{year}{2021}\natexlab{}.
\newblock \showarticletitle{Segmenter: Transformer for semantic segmentation}. In \bibinfo{booktitle}{\emph{Proceedings of the IEEE/CVF international conference on computer vision}}. \bibinfo{pages}{7262--7272}.
\newblock


\bibitem[Tian et~al\mbox{.}(2021)]%
        {tian2021good}
\bibfield{author}{\bibinfo{person}{Yu Tian}, \bibinfo{person}{Jian Ren}, \bibinfo{person}{Menglei Chai}, \bibinfo{person}{Kyle Olszewski}, \bibinfo{person}{Xi Peng}, \bibinfo{person}{Dimitris~N Metaxas}, {and} \bibinfo{person}{Sergey Tulyakov}.} \bibinfo{year}{2021}\natexlab{}.
\newblock \showarticletitle{A good image generator is what you need for high-resolution video synthesis}.
\newblock \bibinfo{journal}{\emph{arXiv preprint arXiv:2104.15069}} (\bibinfo{year}{2021}).
\newblock


\bibitem[Touvron et~al\mbox{.}(2021)]%
        {touvron2021training}
\bibfield{author}{\bibinfo{person}{Hugo Touvron}, \bibinfo{person}{Matthieu Cord}, \bibinfo{person}{Matthijs Douze}, \bibinfo{person}{Francisco Massa}, \bibinfo{person}{Alexandre Sablayrolles}, {and} \bibinfo{person}{Herv{\'e} J{\'e}gou}.} \bibinfo{year}{2021}\natexlab{}.
\newblock \showarticletitle{Training data-efficient image transformers \& distillation through attention}. In \bibinfo{booktitle}{\emph{International conference on machine learning}}. PMLR, \bibinfo{pages}{10347--10357}.
\newblock


\bibitem[Tsuchida et~al\mbox{.}(2019)]%
        {aist-dance-db}
\bibfield{author}{\bibinfo{person}{Shuhei Tsuchida}, \bibinfo{person}{Satoru Fukayama}, \bibinfo{person}{Masahiro Hamasaki}, {and} \bibinfo{person}{Masataka Goto}.} \bibinfo{year}{2019}\natexlab{}.
\newblock \showarticletitle{AIST Dance Video Database: Multi-genre, Multi-dancer, and Multi-camera Database for Dance Information Processing}. In \bibinfo{booktitle}{\emph{Proceedings of the 20th International Society for Music Information Retrieval Conference, {ISMIR} 2019}}. \bibinfo{address}{Delft, Netherlands}.
\newblock


\bibitem[Twindom(2022)]%
        {twindom}
\bibfield{author}{\bibinfo{person}{Twindom}.} \bibinfo{year}{2022}\natexlab{}.
\newblock \bibinfo{booktitle}{\emph{Twindom 3D Avatar Dataset}}.
\newblock
\urldef\tempurl%
\url{https://web.twindom.com/}
\showURL{%
\tempurl}


\bibitem[Vaswani et~al\mbox{.}(2017)]%
        {vaswani2017attention}
\bibfield{author}{\bibinfo{person}{Ashish Vaswani}, \bibinfo{person}{Noam Shazeer}, \bibinfo{person}{Niki Parmar}, \bibinfo{person}{Jakob Uszkoreit}, \bibinfo{person}{Llion Jones}, \bibinfo{person}{Aidan~N Gomez}, \bibinfo{person}{{\L}ukasz Kaiser}, {and} \bibinfo{person}{Illia Polosukhin}.} \bibinfo{year}{2017}\natexlab{}.
\newblock \showarticletitle{Attention is all you need}.
\newblock \bibinfo{journal}{\emph{Advances in neural information processing systems}}  \bibinfo{volume}{30} (\bibinfo{year}{2017}).
\newblock


\bibitem[Wang et~al\mbox{.}(2023)]%
        {wang2023disco}
\bibfield{author}{\bibinfo{person}{Tan Wang}, \bibinfo{person}{Linjie Li}, \bibinfo{person}{Kevin Lin}, \bibinfo{person}{Chung-Ching Lin}, \bibinfo{person}{Zhengyuan Yang}, \bibinfo{person}{Hanwang Zhang}, \bibinfo{person}{Zicheng Liu}, {and} \bibinfo{person}{Lijuan Wang}.} \bibinfo{year}{2023}\natexlab{}.
\newblock \showarticletitle{Disco: Disentangled control for referring human dance generation in real world}.
\newblock \bibinfo{journal}{\emph{arXiv e-prints}} (\bibinfo{year}{2023}), \bibinfo{pages}{arXiv--2307}.
\newblock


\bibitem[Wang et~al\mbox{.}(2021)]%
        {wang2021one}
\bibfield{author}{\bibinfo{person}{Ting-Chun Wang}, \bibinfo{person}{Arun Mallya}, {and} \bibinfo{person}{Ming-Yu Liu}.} \bibinfo{year}{2021}\natexlab{}.
\newblock \showarticletitle{One-shot free-view neural talking-head synthesis for video conferencing}. In \bibinfo{booktitle}{\emph{Proceedings of the IEEE/CVF conference on computer vision and pattern recognition}}. \bibinfo{pages}{10039--10049}.
\newblock


\bibitem[Wang et~al\mbox{.}(2020)]%
        {wang2020g3an}
\bibfield{author}{\bibinfo{person}{Yaohui Wang}, \bibinfo{person}{Piotr Bilinski}, \bibinfo{person}{Francois Bremond}, {and} \bibinfo{person}{Antitza Dantcheva}.} \bibinfo{year}{2020}\natexlab{}.
\newblock \showarticletitle{G3AN: Disentangling appearance and motion for video generation}. In \bibinfo{booktitle}{\emph{Proceedings of the IEEE/CVF Conference on Computer Vision and Pattern Recognition}}. \bibinfo{pages}{5264--5273}.
\newblock


\bibitem[Xie et~al\mbox{.}(2021)]%
        {xie2021segformer}
\bibfield{author}{\bibinfo{person}{Enze Xie}, \bibinfo{person}{Wenhai Wang}, \bibinfo{person}{Zhiding Yu}, \bibinfo{person}{Anima Anandkumar}, \bibinfo{person}{Jose~M Alvarez}, {and} \bibinfo{person}{Ping Luo}.} \bibinfo{year}{2021}\natexlab{}.
\newblock \showarticletitle{SegFormer: Simple and efficient design for semantic segmentation with transformers}.
\newblock \bibinfo{journal}{\emph{Advances in neural information processing systems}}  \bibinfo{volume}{34} (\bibinfo{year}{2021}), \bibinfo{pages}{12077--12090}.
\newblock


\bibitem[Xu et~al\mbox{.}(2023)]%
        {xu2023magicanimate}
\bibfield{author}{\bibinfo{person}{Zhongcong Xu}, \bibinfo{person}{Jianfeng Zhang}, \bibinfo{person}{Jun~Hao Liew}, \bibinfo{person}{Hanshu Yan}, \bibinfo{person}{Jia-Wei Liu}, \bibinfo{person}{Chenxu Zhang}, \bibinfo{person}{Jiashi Feng}, {and} \bibinfo{person}{Mike~Zheng Shou}.} \bibinfo{year}{2023}\natexlab{}.
\newblock \showarticletitle{Magicanimate: Temporally consistent human image animation using diffusion model}.
\newblock \bibinfo{journal}{\emph{arXiv preprint arXiv:2311.16498}} (\bibinfo{year}{2023}).
\newblock


\bibitem[Yang et~al\mbox{.}(2024)]%
        {yang2024direct}
\bibfield{author}{\bibinfo{person}{Shiyuan Yang}, \bibinfo{person}{Liang Hou}, \bibinfo{person}{Haibin Huang}, \bibinfo{person}{Chongyang Ma}, \bibinfo{person}{Pengfei Wan}, \bibinfo{person}{Di Zhang}, \bibinfo{person}{Xiaodong Chen}, {and} \bibinfo{person}{Jing Liao}.} \bibinfo{year}{2024}\natexlab{}.
\newblock \showarticletitle{Direct-a-Video: Customized Video Generation with User-Directed Camera Movement and Object Motion}.
\newblock \bibinfo{journal}{\emph{arXiv preprint arXiv:2402.03162}} (\bibinfo{year}{2024}).
\newblock


\bibitem[Yi et~al\mbox{.}(2023)]%
        {yi2023generating}
\bibfield{author}{\bibinfo{person}{Hongwei Yi}, \bibinfo{person}{Hualin Liang}, \bibinfo{person}{Yifei Liu}, \bibinfo{person}{Qiong Cao}, \bibinfo{person}{Yandong Wen}, \bibinfo{person}{Timo Bolkart}, \bibinfo{person}{Dacheng Tao}, {and} \bibinfo{person}{Michael~J Black}.} \bibinfo{year}{2023}\natexlab{}.
\newblock \showarticletitle{Generating Holistic 3D Human Motion from Speech}. In \bibinfo{booktitle}{\emph{IEEE Conference on Computer Vision and Pattern Recognition (CVPR)}}. \bibinfo{pages}{469--480}.
\newblock


\bibitem[Yu et~al\mbox{.}(2021)]%
        {tao2021function4d}
\bibfield{author}{\bibinfo{person}{Tao Yu}, \bibinfo{person}{Zerong Zheng}, \bibinfo{person}{Kaiwen Guo}, \bibinfo{person}{Pengpeng Liu}, \bibinfo{person}{Qionghai Dai}, {and} \bibinfo{person}{Yebin Liu}.} \bibinfo{year}{2021}\natexlab{}.
\newblock \showarticletitle{Function4D: Real-time Human Volumetric Capture from Very Sparse Consumer RGBD Sensors}. In \bibinfo{booktitle}{\emph{IEEE Conference on Computer Vision and Pattern Recognition (CVPR2021)}}.
\newblock


\bibitem[Yuan et~al\mbox{.}(2021)]%
        {yuan2021tokens}
\bibfield{author}{\bibinfo{person}{Li Yuan}, \bibinfo{person}{Yunpeng Chen}, \bibinfo{person}{Tao Wang}, \bibinfo{person}{Weihao Yu}, \bibinfo{person}{Yujun Shi}, \bibinfo{person}{Zi-Hang Jiang}, \bibinfo{person}{Francis~EH Tay}, \bibinfo{person}{Jiashi Feng}, {and} \bibinfo{person}{Shuicheng Yan}.} \bibinfo{year}{2021}\natexlab{}.
\newblock \showarticletitle{Tokens-to-token vit: Training vision transformers from scratch on imagenet}. In \bibinfo{booktitle}{\emph{Proceedings of the IEEE/CVF international conference on computer vision}}. \bibinfo{pages}{558--567}.
\newblock


\bibitem[Zhang et~al\mbox{.}(2023a)]%
        {zhang2023closet}
\bibfield{author}{\bibinfo{person}{Hongwen Zhang}, \bibinfo{person}{Siyou Lin}, \bibinfo{person}{Ruizhi Shao}, \bibinfo{person}{Yuxiang Zhang}, \bibinfo{person}{Zerong Zheng}, \bibinfo{person}{Han Huang}, \bibinfo{person}{Yandong Guo}, {and} \bibinfo{person}{Yebin Liu}.} \bibinfo{year}{2023}\natexlab{a}.
\newblock \showarticletitle{Closet: Modeling clothed humans on continuous surface with explicit template decomposition}. In \bibinfo{booktitle}{\emph{Proceedings of the IEEE/CVF Conference on Computer Vision and Pattern Recognition}}. \bibinfo{pages}{501--511}.
\newblock


\bibitem[Zhang et~al\mbox{.}(2023b)]%
        {zhang2023adding}
\bibfield{author}{\bibinfo{person}{Lvmin Zhang}, \bibinfo{person}{Anyi Rao}, {and} \bibinfo{person}{Maneesh Agrawala}.} \bibinfo{year}{2023}\natexlab{b}.
\newblock \showarticletitle{Adding conditional control to text-to-image diffusion models}. In \bibinfo{booktitle}{\emph{Proceedings of the IEEE/CVF International Conference on Computer Vision}}. \bibinfo{pages}{3836--3847}.
\newblock


\bibitem[Zhang et~al\mbox{.}(2021)]%
        {lightcap2021}
\bibfield{author}{\bibinfo{person}{Yuxiang Zhang}, \bibinfo{person}{Zhe Li}, \bibinfo{person}{Liang An}, \bibinfo{person}{Mengcheng Li}, \bibinfo{person}{Tao Yu}, {and} \bibinfo{person}{Yebin Liu}.} \bibinfo{year}{2021}\natexlab{}.
\newblock \showarticletitle{Light-weight Multi-person Total Capture Using Sparse Multi-view Cameras}. In \bibinfo{booktitle}{\emph{IEEE International Conference on Computer Vision}}.
\newblock


\bibitem[Zheng et~al\mbox{.}(2023)]%
        {zheng2023fast}
\bibfield{author}{\bibinfo{person}{Hongkai Zheng}, \bibinfo{person}{Weili Nie}, \bibinfo{person}{Arash Vahdat}, {and} \bibinfo{person}{Anima Anandkumar}.} \bibinfo{year}{2023}\natexlab{}.
\newblock \showarticletitle{Fast training of diffusion models with masked transformers}.
\newblock \bibinfo{journal}{\emph{arXiv preprint arXiv:2306.09305}} (\bibinfo{year}{2023}).
\newblock


\bibitem[Zheng et~al\mbox{.}(2021)]%
        {zheng2021rethinking}
\bibfield{author}{\bibinfo{person}{Sixiao Zheng}, \bibinfo{person}{Jiachen Lu}, \bibinfo{person}{Hengshuang Zhao}, \bibinfo{person}{Xiatian Zhu}, \bibinfo{person}{Zekun Luo}, \bibinfo{person}{Yabiao Wang}, \bibinfo{person}{Yanwei Fu}, \bibinfo{person}{Jianfeng Feng}, \bibinfo{person}{Tao Xiang}, \bibinfo{person}{Philip~HS Torr}, {et~al\mbox{.}}} \bibinfo{year}{2021}\natexlab{}.
\newblock \showarticletitle{Rethinking semantic segmentation from a sequence-to-sequence perspective with transformers}. In \bibinfo{booktitle}{\emph{Proceedings of the IEEE/CVF conference on computer vision and pattern recognition}}. \bibinfo{pages}{6881--6890}.
\newblock


\bibitem[Zhu et~al\mbox{.}(2024)]%
        {zhu2024champ}
\bibfield{author}{\bibinfo{person}{Shenhao Zhu}, \bibinfo{person}{Junming~Leo Chen}, \bibinfo{person}{Zuozhuo Dai}, \bibinfo{person}{Yinghui Xu}, \bibinfo{person}{Xun Cao}, \bibinfo{person}{Yao Yao}, \bibinfo{person}{Hao Zhu}, {and} \bibinfo{person}{Siyu Zhu}.} \bibinfo{year}{2024}\natexlab{}.
\newblock \showarticletitle{Champ: Controllable and Consistent Human Image Animation with 3D Parametric Guidance}.
\newblock \bibinfo{journal}{\emph{arXiv preprint arXiv:2403.14781}} (\bibinfo{year}{2024}).
\newblock


\end{thebibliography}

\clearpage

\begin{figure*}
    \centering
    \includegraphics[width=\linewidth]{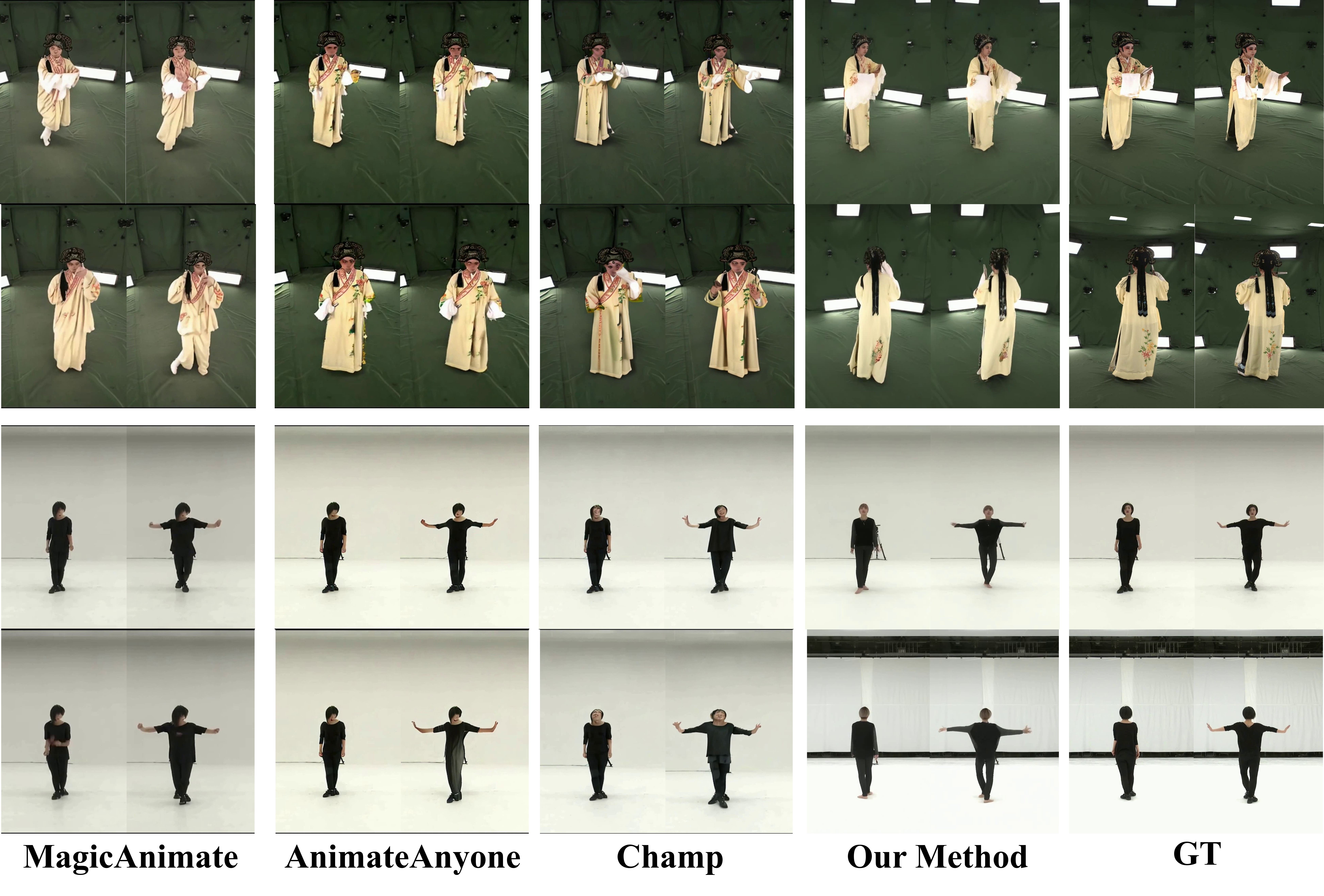}
    \caption{Qualitative comparison multi-view video.}
    \label{fig:mv_comp}
\end{figure*}

\begin{figure*}
    \centering
    \includegraphics[width=0.9\linewidth]{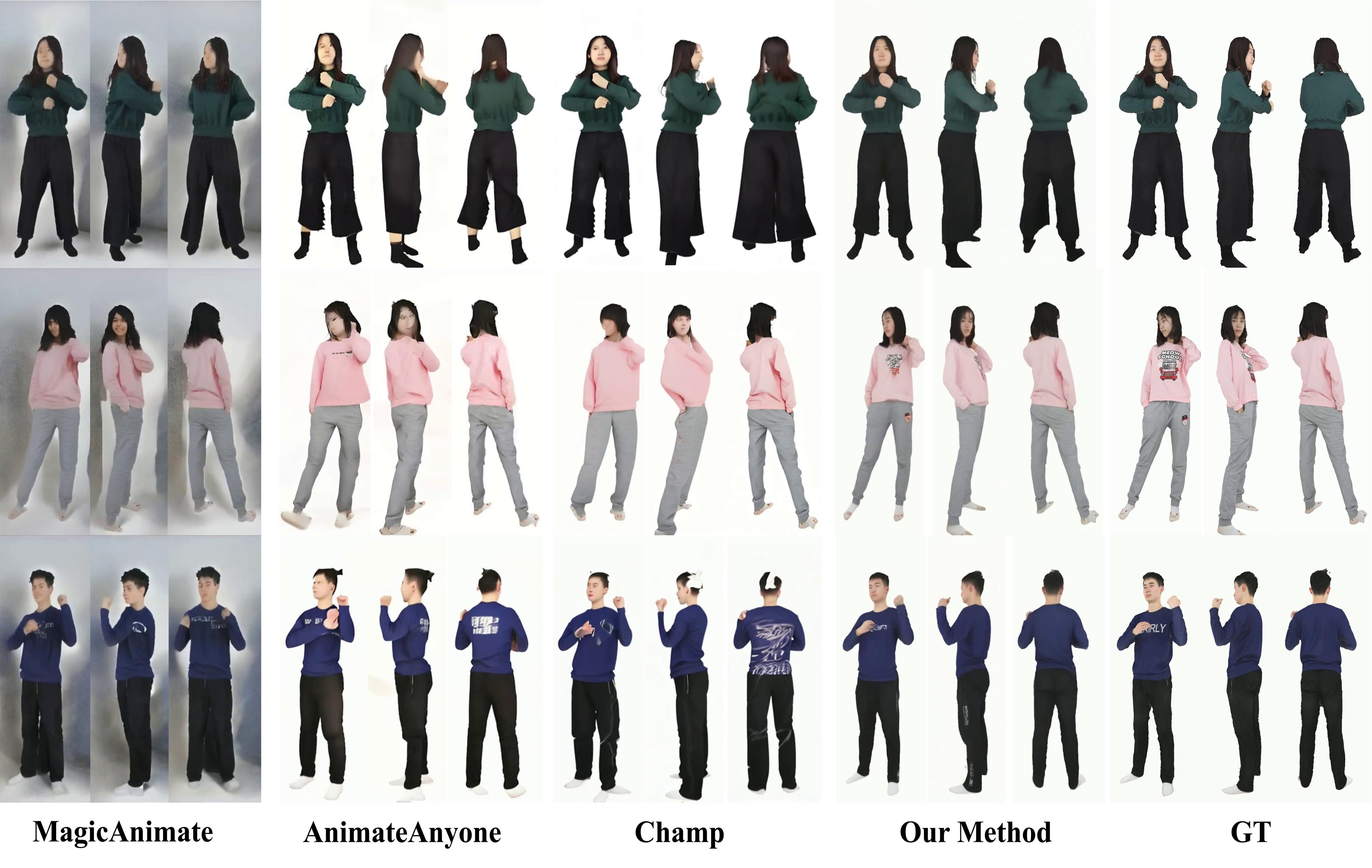}
    \caption{Qualitative comparison 3D static  video.}
    \label{fig:3d_comp}
\end{figure*}

\begin{figure*}
    \centering
    \includegraphics[width=0.9\linewidth]{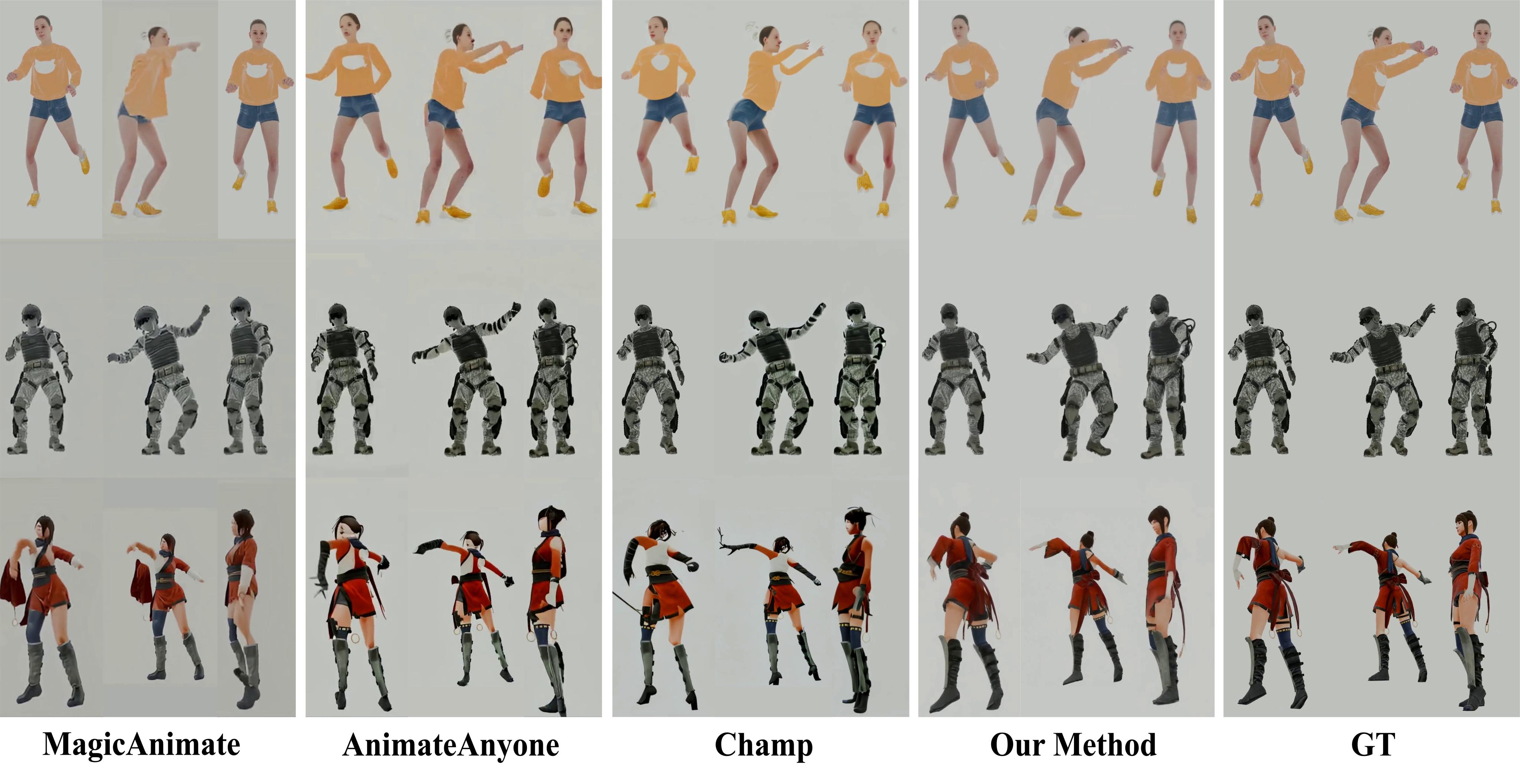}
    \caption{Qualitative comparison free-view  video.}
    \label{fig:4d_comp}
\end{figure*}

\end{document}